%% file: main.tex
\documentclass[runningheads]{llncs}

\usepackage{eccv}

\input{preamble}

\usepackage{eccvabbrv}

\usepackage{graphicx}
\usepackage{booktabs}

\usepackage[accsupp]{axessibility}  

\usepackage{hyperref}

\usepackage{orcidlink}

\begin{document}

\title{Quality and Quantity: Unveiling a Million High-Quality Images for Text-to-Image Synthesis in Fashion Design}
\titlerunning{Quality and Quantity}


\author{
\centerline{Jia Yu$^{*,1,2}$ ~~ Lichao Zhang$^{*,2}$~~ Zijie Chen$^{*,1,2}$ ~~ Fayu Pan$^{3}$ ~~ Miaomiao Wen$^{4}$~~ Yuming Yan$^{3}$}
\centerline{~~ Fangsheng Weng$^{3}$~~ Shuai Zhang$^{1,2}$ ~~ Lili Pan$^{\dagger,5}$ ~~ Zhenzhong Lan$^{\dagger,2}$}
\centerline{\normalfont{$^1$Zhejiang University} \quad {$^2$Westlake University}\quad $^{3}$Westlake Xinchen  Technology Co. Ltd} 
\centerline{$^{4}$Zhiyi Tech \quad $^5$University of Electronic Science and Technology of China}
\centerline{\texttt{yujia@westlake.edu.cn}}
}

\authorrunning{J.~Yu et al.}

\institute{}


\maketitle 

\input{sec/0_abstract}    
\input{sec/1_intro}

\input{sec/2_rw}
\input{sec/3_data}

\input{sec/4_ana}
\input{sec/5_exp}

\input{sec/6_con}

\clearpage
\bibliographystyle{splncs04}
\bibliography{main}

\input{sec/X_suppl}

\end{document}

%% file: preamble.tex
\usepackage[dvipsnames]{xcolor}
\usepackage{graphicx}
\usepackage{amsmath}
\usepackage{amssymb}
\usepackage{booktabs}
\usepackage{multirow}
\usepackage{utfsym}
\usepackage{diagbox}
\usepackage{slashbox}
\usepackage{float}
\usepackage{placeins}
\usepackage{caption}
\usepackage{stfloats}
\usepackage[bottom]{footmisc} 

\iftrue
\fi

\usepackage{tablefootnote}

%% file: sec/0_abstract.tex
\begin{abstract}
The fusion of AI and fashion design has emerged as a promising research area. However, the lack of extensive, interrelated data used for training fashion models has hindered the full potential of AI in this area. To address this problem, we present the Fashion-Diffusion dataset, a product of multiple years' rigorous effort. This dataset comprises over a million high-quality fashion images, paired with detailed text descriptions. Sourced from a diverse range of geographical locations and cultural backgrounds, the dataset encapsulates global fashion trends. The images have been meticulously annotated with fine-grained attributes related to clothing and humans, simplifying the fashion design process into a Text-to-Image (T2I) task. The Fashion-Diffusion dataset not only provides high-quality text-image pairs and diverse human-garment pairs but also serves as a large-scale resource about humans, thereby facilitating research in T2I generation. Moreover, to foster standardization in the T2I-based fashion design field, we propose a new benchmark comprising multiple subsets for evaluating the performance of fashion design models. 
Experimental results illustrate our dataset's superiority in both quality (FID: 8.33 vs 15.32, IS: 6.95 vs 4.7, CLIPScore: 0.83 vs 0.70) and quantity (1.04M fashion images at a 768x1152 resolution). This sets a new benchmark for future research in fashion design.

\end{abstract}

%% file: sec/1_intro.tex
\section{Introduction}
\label{sec:intro}

In the past couple of years, artificial intelligence generated content (AIGC) technology has achieved tremendous success and shows the potential to revolutionize various industries and improve human experiences~\cite{bert,gpt3,chatgpt,dall-e2,llama,llama2,flamingo, Rombach_2022_CVPR, zhu2023tryondiffusion}. 
With the advent of text-to-image generation models like DALL-E~\cite{dall-e2}, Stable Diffusion~\cite{Rombach_2022_CVPR} and Imagen~\cite{zhu2023tryondiffusion}, people are starting to believe that AI creation or design is on the cusp of becoming a reality.

\begin{figure*}
    \centering
    \includegraphics[width=\linewidth]{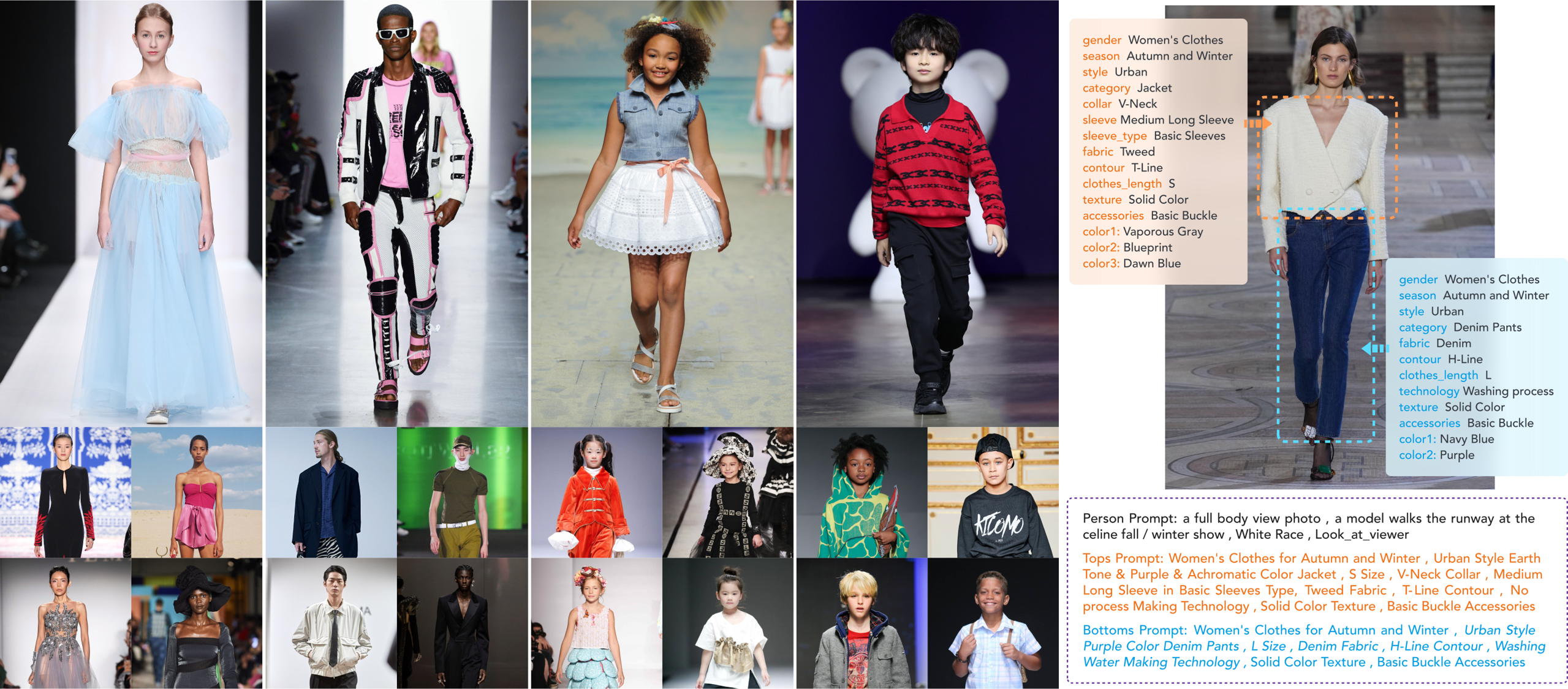}
    \caption{Overview of Fashion-Diffusion. Our Fashion-Diffusion Dataset contains 1,044,491 high-resolution, high-quality fashion images with 1,593,808 high-quality text descriptions, which include descriptions about both garments and humans.}
    \label{fig: overview}
\end{figure*}


The intersection of artificial intelligence (AI) and fashion design has recently garnered significant interest within the realm of computer vision~\cite{lewis2021tryongan, zhu2023tryondiffusion, zhu2017your, zhang2023diffcloth,zhang2022armani,jiang2022text2human,sun2023sgdiff,morelli2022dress}.
A primary obstacle hindering the development of AI fashion design is the lack of a vast, high-quality image dataset paired with abundant text descriptions.
Several existing datasets, such as Prada \cite{zhu2017your} and DeepFashion-MM \cite{jiang2022text2human}, contain a relatively small number of fashion images, with fewer than 100,000 images, and lack comprehensive textual descriptions concerning fine-grained attributes paired with fashion image. 
On the other hand, some datasets, such as DeepFashion \cite{liu2016deepfashion} and CM-Fashion \cite{zhang2022armani}, surpass the aforementioned datasets in scale; however, images in these datasets either have restricted image resolution (\emph{e.g.}, $256\times256$ for Deepfashion) or only comprise half-body or individual garments. 
Limitations in both the quantity and quality of datasets may weaken the capability of fashion design models trained on them.

Creating a vast text-image fashion dataset also with high-quality presents a formidable challenge due to several factors. The initial hurdle is the daunting process of collecting a large set of high-quality images with paired text descriptions that exhibit sufficient diversity. Additionally, ensuring that the fashion images incorporate human figures and that the texts provide detailed human descriptions further adds to the data collection burden. Finally, annotating this dataset with intricate clothing attributes is also non-trivial, in which manual annotation of images with detailed attributes is required.

To overcome the above challenges, we have dedicated several years to collecting a large and high-quality fashion dataset called Fashion-Diffusion. 
Launched in 2018, our Fashion-Diffusion dataset efforts consist of collecting and carefully curating fashion images sourced from a vast collection of high-quality clothing images. These images, sourced from a wide range of geographical locations and cultural contexts, encapsulate global fashion trends. For the construction of Fashion-Diffusion, we employed a blend of manual and automated annotation techniques for subject detection and classification. In collaboration with clothing design experts, we identified a set of clothing-related attributes, including some that are particularly detailed, resulting in a total of {8037} labeled attributes. Finally, we amalgamated and augmented the information from the initial stages, using BLIP~\cite{li2022blip} for caption generation, followed by manual review and correction of the produced captions.

The Fashion-Diffusion dataset holds distinct advantages over its predecessors. Firstly, it offers high-quality text-image pairs: the images in the Fashion-Diffusion dataset have a resolution of $768 \times 1152$, ensuring a high level of detail for analysis (see Fig.~\ref{fig: overview}). The text prompts about humans and clothing are also detailed, with lengths of $15\sim25$ words and $35\sim55$ words respectively, a level of detail seldom found in other datasets. The relevance between image and text in Fashion-Diffusion is superior, boasting a CLIPScore of 0.80. Secondly, the dataset contains an extensive number of fashion images (1,044,491), spanning {8037} attributes for clothing and humans. These features simplify the fashion design process into a Text-to-Image (T2I) task, eliminating the need for auxiliary input in other forms.
Finally, the dataset offers diverse garment-human pairs encompassing persons of all races and ages, wearing garments of 52 fine-gained categories. The contributions of this work can be summarized as follows:


\begin{itemize}
\item We have compiled the Fashion-Diffusion dataset, which includes 1,044,491 high-quality fashion images with a resolution of $768\times1152$, each with detailed text descriptions sourced from 8037 attributes. This dataset is the first to provide over a million fashion images comprising both garments and humans. This dataset will aid the research in fashion and be made public upon paper acceptance. Beyond being a large-scale fashion dataset, Fashion-Diffusion is also a large-scale dataset of human images providing detailed clothing-related attributes. These features will also be instrumental in advancing research on T2I generation.
\item We have conducted a thorough statistical analysis of the Fashion-Diffusion dataset, showing it includes high-quality text-image and diverse human-garment pairs.
\item We propose a novel benchmark for assessing the efficacy of fashion design models, promoting the standardization within the T2I-based fashion design domain.
\end{itemize}

\begin{table}[!hbt]
  \centering
  \small
  \begin{tabular}{|p{.24\columnwidth}|p{.075\columnwidth}|p{.14\columnwidth}|p{.065\columnwidth}|p{.07\columnwidth}|p{.065\columnwidth}|p{.06\columnwidth}|p{.07\columnwidth}|p{.045\columnwidth}|p{.065\columnwidth}|}
      \hline
      \multirow{2}{*}{Dataset}  & \multirow{2}{*}{Size}& \multirow{2}{*}{Resolution} & \multicolumn{2}{c|}{Text Caption} &\multicolumn{2}{c|}{Person}& Garm.& \multirow{2}{*}{Cls.}&\multirow{2}{*}{Attrs.}\\ \cline{4-8} 
      &&&Exist.&Length&Exist. &Age& Cat. & & \\ \hline
   {Clothing-Attrs.}\cite{chen2012describing} & 1.8K & $< 750\times750$ & $\times$ &-& \checkmark & adult& & 11 &26\\ \hline      
      ACWS\cite{bossard2013apparel} & 145K & $< 270\times270$ & $\times$ &-& part.  & all& & 8 & 78 \\ \hline        DeepFashion\cite{liu2016deepfashion}&800K&$256\times256$&$\times$&-&\checkmark& adult&50& 5&1000\\ \hline
      Prada\cite{zhu2017your} & 78K & $256\times256$ & \checkmark & 8.07 & \checkmark & adult& & -& -\\ \hline
    \scriptsize{DeepFashion-MM\cite{jiang2022text2human}} &44K&$512\times1024$&\checkmark&40.44 & \checkmark& adult& & &28\\ \hline
      Dress Code\cite{morelli2022dress} &107K&$768\times1024$&$\times$&-&part.&adult& & &- \\ \hline
      CM-Fashion\cite{zhang2022armani} & 500K & - & \checkmark &- & $\times$  & -& & &- \\ \hline
      SG-Fashion\cite{sun2023sgdiff} & 17K & - & \checkmark &-& $\times$ & -&  & &-\\   \hline  
      FIRST\cite{huang2023first} & 1.00M &  512$\times$512 & \checkmark &- & \checkmark  & -& & &- \\
      \hline\hline
      \textbf{Fashion-Diffusion} &\textbf{1.04}M & \textbf{768$\times$1152}  & \checkmark &\textbf{67.45}& \checkmark &all&\textbf{52} & \textbf{23} &\textbf{8037}\\ \hline
      \end{tabular}
  \caption{Statistics of Fashion-Diffusion dataset and its comparison with existing public fashion image datasets. Fashion-Diffusion dataset consists of high-resolution fashion image dataset containing over 1.04M text-image pairs of full-body people in all ages and genders, dressed in extremely diverse garments in 23 classes with 8037 fine-grained annotated attributes. `Exist.', `Garm.', `Cat.', `Cls', `Attrs.', `part.' are the abbreviations of `Existence', `Garmnet', `Category', `Class', `Attributes' and `partial' respectively.}
\label{table:datasets comparison}
\end{table}

%% file: sec/2_rw.tex
\section{Related Work}
\label{sec:rw}

In this section, we will review the fashion text-image datasets that are utilized in text-to-image generation model training. Then we sort out attractive text-image generation-models.

\subsection{Fashion Image Datasets}
Fashion image datasets serves various downstream tasks, including virtual try-on, image-to-image translation, image retrieval, demonstrating their significance in both academic research and industrial applications. However, due to commercial reasons most fashion image datasets\cite{chen2015deep}\cite{choi2021viton}\cite{hadi2015buy}\cite{han2018viton}\cite{huang2015cross}\cite{lewis2021tryongan} are not publicly available.

To the best of our knowledge, Clothing Attributes Dataset\cite{chen2012describing} is the first fashion image dataset available to the public. It includes 1,856 images of clothed people, with 7 categories of garments and 26 other attributes annotated using SVM and CRF.
ACWS\cite{bossard2013apparel} is a 145K fashion image dataset but is low in image resolution, and not all images in it contain humans. Garments appear in ACWS fall in 15 categories and are annotated with 78 attributes.
DeepFashion\cite{liu2016deepfashion} is a large-scale fashion image dataset of 800K images (with a resolution of $256\times256$) of dressed humans. It includes clothes from 50 categories annotated in by 1000 attributes. The images are also annotated with landmarks to locate the garments. These early fashion image datasets do not include text captions, probably due to the deficiency of cross-modal learning and NLP at that time.
This limitation impedes the use of DeepFashion for training current T2I models in fashion design.

More recent fashion image datasets began to include text captions.
A subset of 78K images from DeepFashion dataset is collected by \cite{zhu2017your} and manually annotated using a short sentence each image. They adopt landmark annotations from DeepFashion.
DeepFashion-MM\cite{jiang2022text2human} is a dataset containing 44K human images, each with a textual description along with human parsing and dense pose features. DeepFashion-MM categorized garments in images into 23 categories and further annotated 28 attributes for the garments. 
The above two datasets contain a relatively small number of fashion images, both with fewer than 100,000 images. 
CM-Fashion\cite{zhang2022armani} and SG-Fashion\cite{sun2023sgdiff} are fashion clothes datasets with no human in images. Both datasets include text captions and are supposed to be public, but not yet now.

Previous fashion image datasets often include additional visual features such as dense pose, landmark, human parsing, etc. Such visual features are designed to simplify the tasks for outdated neural networks. 
The advancement of diffusion models \cite{ho2020denoising} \cite{song2020denoising} \cite{Rombach_2022_CVPR} \cite{saharia2022photorealistic} and vision-language models \cite{alayrac2022flamingo}\cite{radford2021learning} \cite{li2022blip} \cite{li2023blip} shows unprecedented ability of high-quality text-to-image generation and understanding cross-modal semantics. We claim that the additional visual features are no longer essential for today's models.
Concurrent to our work, Huang et al. \cite{huang2023first} have introduced a dataset of one million images annotated with texture descriptions for fashion design, known as the FIRST dataset. However, images within the FIRST dataset notably exhibit a lower resolution of $512 \times 512$. Importantly, the attribute descriptions pertaining to fashion design remain undisclosed in their publication, and as of yet, the dataset has not been made publicly accessible.

Table \ref{table:datasets comparison} shows existing public fashion image datasets and their comparison with our dataset. Our dataset is the first public large-scale high-resolution fashion image dataset containing 1.04M text-image pairs of full-body people in all ages and genders dressed in extremely diverse garments, with 8037 fine-grained annotated attributes.



\subsection{Garment synthesis}




For garment synthesis, multiple modalities, e.g. text, mask and pose, are used as the input for generating clothes. 
Text2Human~\cite{jiang2022text2human} translates the given human pose to human parsing with texts about cloth shapes, and then more attributes about the cloth textures are used to generate the final human image.  
DiffCloth~\cite{zhang2023diffcloth} uses the parsing solution to segment the text and cloth independently, then matches them together by using bipartite matching, and further strengthens the similarity by aligning cross-attention semantics.

With thorough differences, we do not need any labeled image pairs as the input of generation models. Neither, We do not need auxiliary input in other modalities. We input pure yet exhaustive text prompt, which can precisely control the category and attributes of generated try-on images directly through original text-to-image generation models.


%% file: sec/3_data.tex
\section{Fashion-Diffusion Dataset}
\label{sec:da_co}


High-quality image data serves as the cornerstone of AI advancement in the field of fashion design. We make efforts to carefully construct the Fashion-Diffusion dataset, starting from source crawling, through data annotation, and all the way to final data filtering. 
Inevitably, the dataset collection process is carried out in a human-in-the-loop manner.

\subsection{Data Collection \& Processing}
\noindent \textbf{Collection.} Our data collection involves a wide range of sources and various capturing methods.
We perform distributed web crawling to grasp large-scale fashion-style images based on public fashion websites, including runway and product sources. However, due to quality concerns and potential copyright issues, we excluded product-derived data. This resulted in a final dataset of totally 1.1 million high-quality runway images.
It is worth mentioning that we strictly complied with the relevant regulations and ensured that all the collected images were publicly available and did not infringe any copyrights during the whole process of the dataset construction.

\noindent \textbf{Processing.} We sample high-quality and diverse fashion images from our raw collections. We adopt pre-process filtering to clean the dataset, obtaining three-level subsets, \emph{i.e.}, Subset100K, Subset200K, and Subset1M. 
For Subset1M, the aspect ration and human faces are our primary considerations. Images with inappropriate aspect rations (<=0.5 or >=0.8) and multiple human faces (>=2) are filtered out. Through this kind of filtering, we derive our ultimate largest subset, \emph{i.e.}, 1,044,491 fashion images.

Moreover, we form a customized filtering procedure with five attributes-related filtering rules, by considering constraints based on scale factor, garment features, human characteristics, image attributes and some specific attribute cases. {Please refer to the Table~\ref{tab:filtering_rule} in appendix for the details.}

For constructing Subset100K, we filtered the collected 1.1 million images using the five rules. We also ensure to preserve the datasets at each stage of the filtering process. This allows us to track the progression and impact of each rule on the final dataset.
Then, we augment our existing Subset100K to reach a total of 200K images. This is accomplished by incorporating approximately 100K new images from our stored data source during the fourth stage, which is after the application of the first four specific filtering rules.
The specific numbers of prompts and images for each subset are clearly listed in Sec.~\ref{quality_res}.

\subsection{Data Annotation}

\begin{figure}[!hbt]
    \centering
    \includegraphics[width=\linewidth]{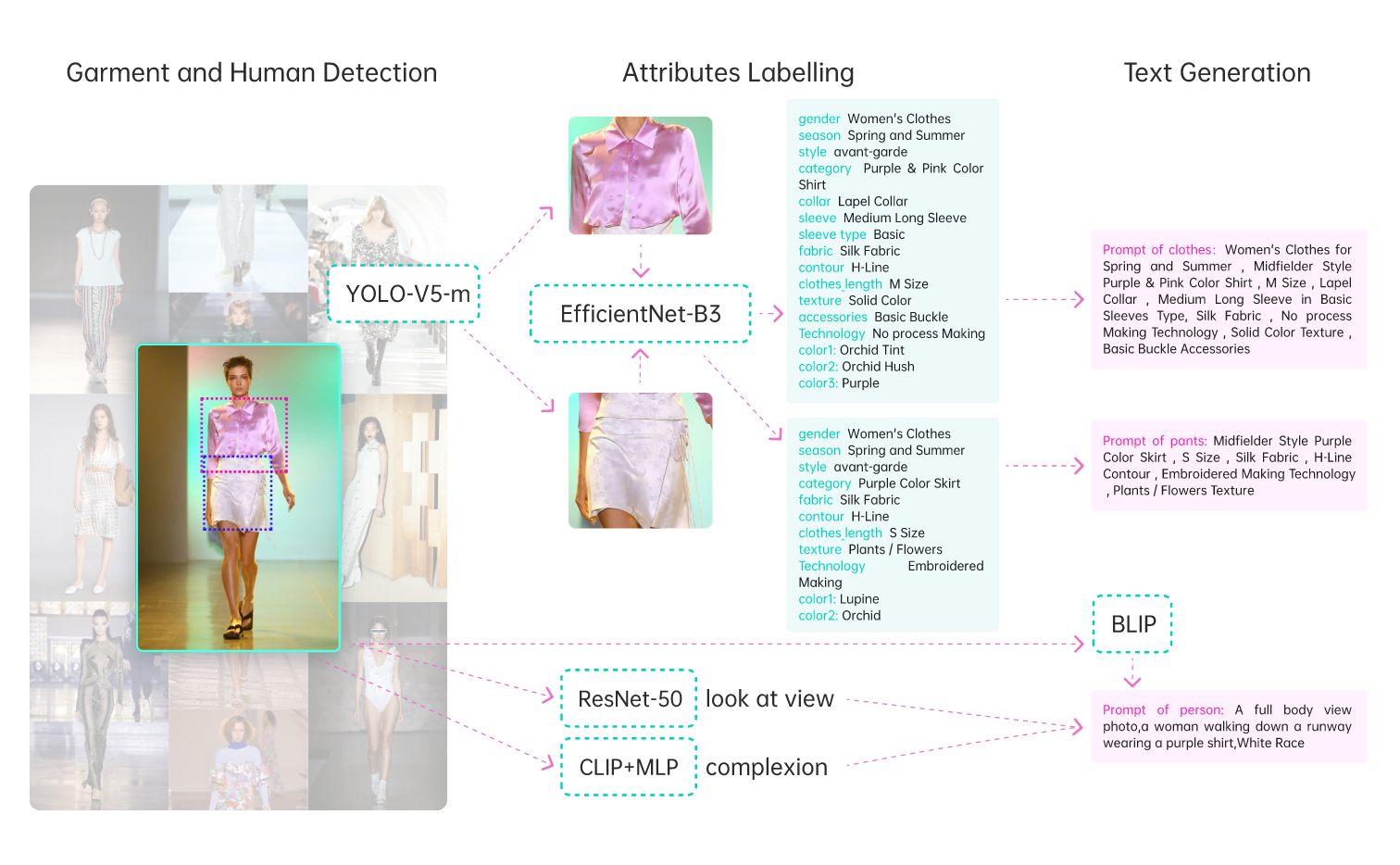}
    \caption{The workflow of the annotation procedure for Fashion-Diffusion. To complete the full annotation task, we employ three stages, namely `Garment and Human Detection', `Attributes Labelling', and `Text Generation', to ensure the annotation in high-quality level as well as the accuracy and professionalism of the text-image information.}
    \label{fig:anno_flow}
\end{figure}

Our primary goal during the data annotation phase is to ensure the accuracy of generated text descriptions. To achieve this goal, we employ a three-stage annotation approach, including garment and human detection, attributes labelling, and text generation. This process is shown in Fig.~\ref{fig:anno_flow}.
In the Garment and human detection stage, we focus on detecting the garment part and humans in the fashion image. In the attributes labeling stage, a classification model is further used to identify attributes of garments or humans. In the text generation stage, we utilize image captioning techniques to produce text descriptions.

\noindent \textbf{Garment and Human Detection.} We employ an efficient object detector,  YOLOv5-m \cite{redmon2016you}, to locate the garment area in fashion images. 
To achieve high-quality annotation, we adopt a hybrid method that includes both manual and automated annotation. Specifically, we first manually annotate a portion of the data, \emph{i.e.}, 400K images with 740K garments. Then, on the labeled data we train detection models. Finally, we accomplish automatic labeling by using well-trained models to detect garments on the remaining images. 
By training on high-quality and extensive manually labeled data, the detector is able to accurately detect the objects, even in images with a variety of background clusters.
We evaluate the models on a validation set, comprising 10\% of the manually labeled dataset (up to 50,000), achieving an accuracy of 0.91, indicating its effectiveness for annotating additional unlabeled data.

\noindent \textbf{Attributes Labelling.} In this stage, we annotate the descriptive attributes related to garments and humans. We employ professionals in the fashion design field to identify 23 classes relevant to fashion design. As in Fig.~\ref{fig:anno_flow}, each class consists of various attributes. Overall, we annotate 8037 attributes about garments and humans. 

We manually annotate partial data across all classes and attributes to train specific classification models, e.g. EfficientNet-B3 model~\cite{tan2019efficientnet}, acting as our labeling classification annotators. The amounts of manually labeled data and corresponding classification accuracy for each class are detailed in Table~\ref{tab:attri} in the appendix. We allocate 10\% of the data for validation, not exceeding 50,000 entries, the similar process as in the stage of human and garment detection, 

Then we use EfficientNet-B3 model~\cite{tan2019efficientnet} finetuned on the manually labeled data to automatically annotate the extra unlabeled data. Based on the detected human in the image, we annotate the image with attributes across classes like gender, garment category, fabric and sleeve type etc.

\noindent \textbf{Text Generation.} Most of our above labeling efforts have been dedicated to describing clothing items. We use ResNet-50 to predict the classes of `look at view' and `view' for the person, and use CLIP+MLP to recognize `complexion' class.
Then, we utilize the BLIP model~\cite{li2022blip} to generate the descriptive text based on the content of the images. 
Finally, we obtain the person prompt by combining captioning descriptions with the above predicted classes.

Finally, we compose a prompt of an image by stacking the person description and the garment description intuitively. Therefore, we can utilize the informative details of both the garment and the person, rather than relying on basic text descriptions found in other fashion datasets~\cite{liu2016deepfashion}~\cite{zhu2017your}, referring to the length of text caption in Table~\ref{table:datasets comparison}.




%% file: sec/4_ana.tex
\section{Statistical Analysis}
\label{sec:ana}


\begin{figure}[t]
    \centering
    \includegraphics[width=\linewidth]{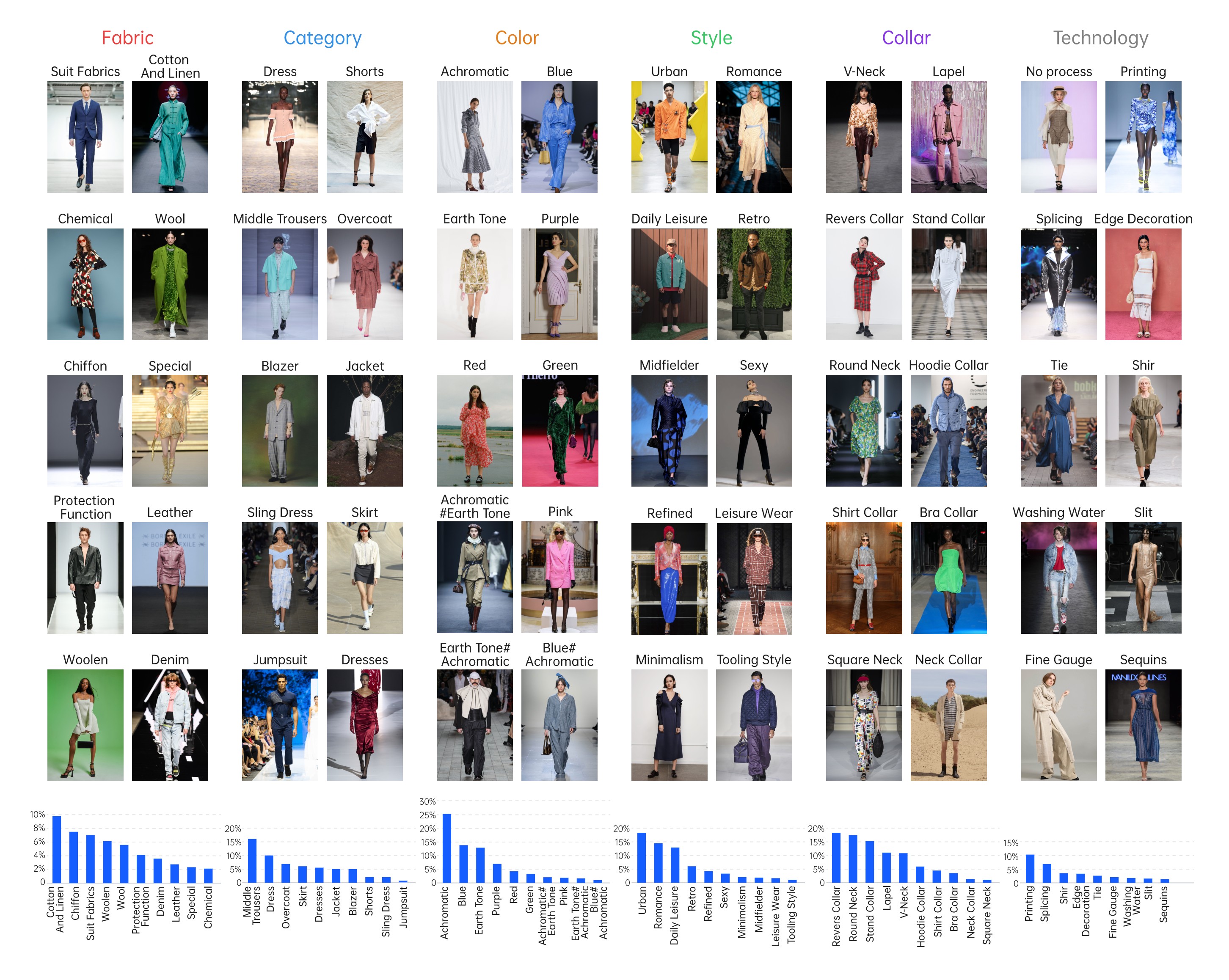}
    \caption{Descriptive attribute distribution with respect to classes of `Fabric', `Category', `Color', `Style', `Collar' and `Technology'. We display exemplar real images for specific attributes under each class and also provide statistics for their top-10 attributes on the bottom row.}
    \label{fig:data_statistic}
\end{figure}

\subsection{Descriptive Attribute Distribution}
We construct the text labels with more than 8K attributes to describe the clothing in a more detailed and professional manner.
Among the 8037 attributes, 6430 pertain to the garment brand.
The text labels in the Fashion-Diffusion dataset cover 23 classes, describing common features and details in fashion design.
Fig.~\ref{fig:data_statistic} shows the distribution of several major attributes under each class, such as `Fabric', `Category', `Color', `Style', `Collar' and `Technology'. These distributions show that the Fashion-Diffusion dataset covers almost all common features in the field of clothing design, indicting the comprehensiveness and specialization of the Fashion-Diffusion dataset in apparel design.

\begin{figure}[!t]
    \centering
    \includegraphics[width=\linewidth]{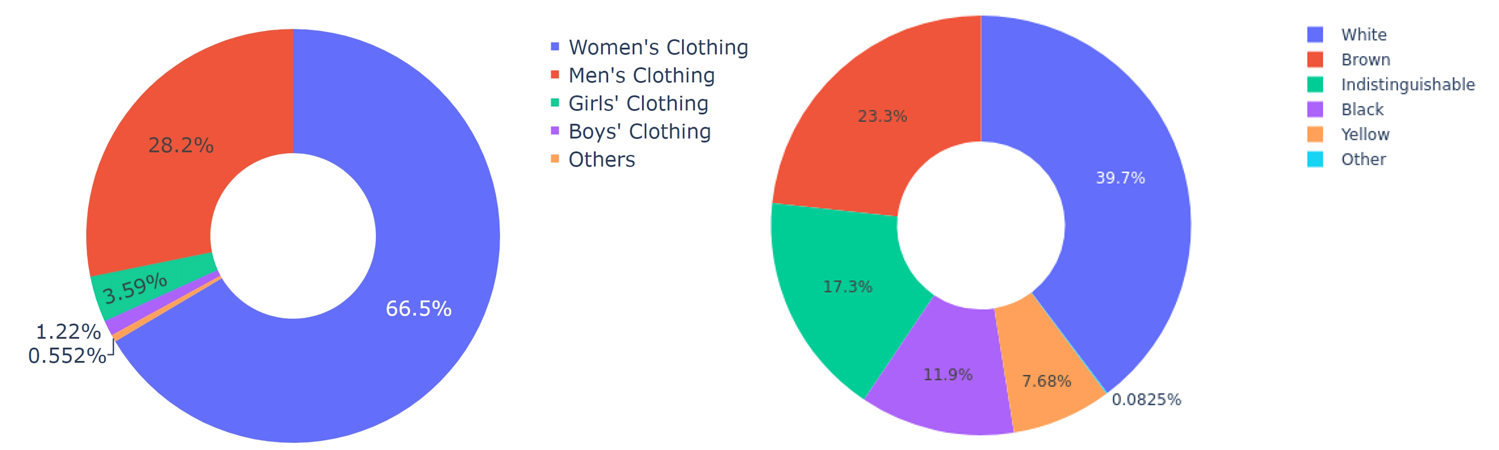}
    \caption{\textbf{Left:} Age and gender distribution in four people factions, i.e., `Women's Clothing', `Men's Clothing', `Girls' Clothing', and `Boys' Clothing', in Fashion-Diffusion dataset. \textbf{Right:} We collect fashion images from a variety of races with different skin colors, making our data more representative in terms of global diversity.}
    \label{fig:gender}
\end{figure}

Fig.~\ref{fig:gender} illustrates the age and gender distribution of the people in the Fashion-Diffusion dataset. It showcases adult clothing designs dominate the dataset as the distribution of real user needs. Fig.~\ref{fig:gender} highlights that the element of `Women's Clothing' dominates the majority of distributions, by an advantage of 66.5\% in our dataset.
Given the extensive size of our dataset, a minor proportion of children clothing distribution still dominates a significant number of text-image pairs, indicating the high diversity of our dataset. 

\begin{figure}[!b]
    \centering
    \includegraphics[width=\linewidth]{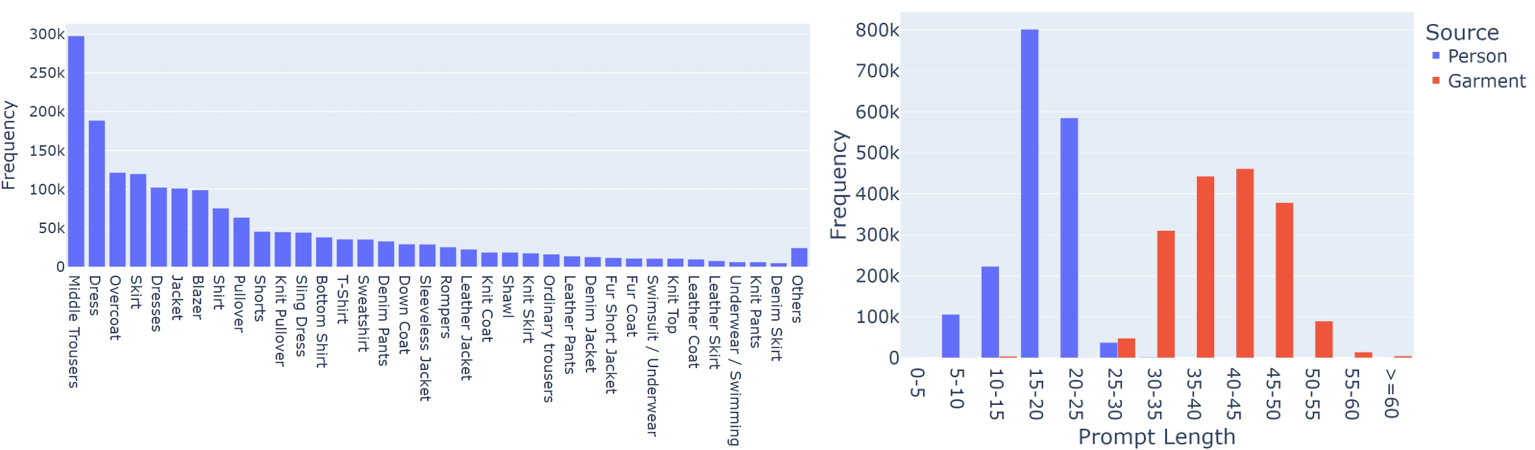}
    \caption{\textbf{Left:} Attributes distribution of the specific `garment category' class describing the type of the clothing in the fashion image. \textbf{Right:} Length distribution of prompts describing both the person and the garment in Fashion-Diffusion dataset.}
    \label{fig:category}
\end{figure}

The left part of Fig.~\ref{fig:category} showcases the distribution of labels for `garment category' class in the Fashion-Diffusion dataset, unveiling a diverse array of prevalent clothing styles. This is rare in other datasets and further confirms the richness and professionalism of the Fashion-Diffusion dataset. For example, we have three attributes, \emph{i.e.}, `fur coat', `leather coat' and `knit coat' in `garment category' class, while there is only a simple `coat' in `category' group in DeepFashion, indicating we have more fine-grained {attributes}.




\subsection{Text-Image Relevance}

As mentioned in Sec.~\ref{sec:da_co}, in the Fashion-Diffusion dataset, the attribute labels of each image are based on the actual features of the image. An effective classification model ensures the accuracy and professionalism of the text in describing image features, and also allows the model to better understand the relationship between text and images. 
The text description for fashion images in Fashion-Diffusion has an average length of $67.45$.
As shown in the right part of Fig.~\ref{fig:category}, the length of the text for describing the person is concentrated in the range of $15\sim25$. Furthermore, the text description for the garment is more detailed and comprehensive, with the length statistically varying from $35\sim55$.

From Table~\ref{tab:T-I Relevance}, we compute the CLIPScore and L2 Distance between the ground-truth texts and images for three datasets, \emph{i.e.}, Prada\cite{zhu2017your}, DeepFashion-MM\cite{jiang2022text2human}.
It showcases that our results generated by human prompt with image are all better than the other datasets. Considering that we need to integrate semantics of both Human and Garment, we sum up the embedding of human prompt and the embedding of garment prompt. Thereby, we use the fused embedding as our identity representation to calculate the CLIPScore and L2 Distance with the image embedding.
These results effectively demonstrate that the Text-Image Relevance in our Fashion-Diffusion dataset is extremely high.


\begin{table}[h]
    \centering
    \scalebox{1}{
    \begin{tabular}{|c|c|c|c|}
    \hline 
    Dataset & Description & CLIPScore$\uparrow$ & L2 Distance$\downarrow$\\ \hline
    \hline
    Prada\cite{zhu2017your} & Prompt &0.65 & 1.21\\ \hline
    DeepFashion-MM\cite{jiang2022text2human} & Prompt & 0.62&1.21 \\ \hline
    \multirow{3}{*}{\textbf{Fashion-Diffusion}} & Human & 0.72 & 1.19\\ \cline{2-4}
    & Garment & 0.62 & 1.23\\ \cline{2-4}
    & Sum &\textbf{0.80} & \textbf{1.17}\\ \hline    
    \end{tabular}
    }
    \caption{Comparisons of Text-Image Relevance between Fashion-Diffusion Dataset and others. We have expressive description texts including both Human and Garment, in contrast, there is one a simple prompt text in compared datasets.}
    \label{tab:T-I Relevance}
\end{table}

%% file: sec/5_exp.tex
\section{Experiments}
\label{sec:exp}

In this section, we present our experimental results to validate the effectiveness of our dataset. It includes quantitative comparisons and qualitative results.


\subsection{Fashion-Diffusion Benchmark}
\label{quality_res}

\textbf{Datasets.}
The Fashion-Diffusion dataset is proposed for training large T2I models in fashion design.
90\% images are randomly chosen for training and 10\% images are used for testing.
To validate our dataset's effectiveness, we split it into three subsets based on image quantity: Subset100K (100K training, 10K testing), Subset200K (200K training, 20K testing), and Subset1M (940K training, 104K testing), as shown in Table~\ref{split_num}.
Furthermore, we use attributes of five commonly used classes `Category' (52), `Style' (25), `Cloth\_len' (3), `Fabric' (26) and `Texture' (33) for specific fine-grained assessment.

\begin{table}[h]
 \centering
 \scalebox{1}{
\begin{tabular}{|c|cc|cc|}
\hline
\multirow{2}{*}{Datasets} & \multicolumn{2}{c}{Training set} & \multicolumn{2}{|c|}{Testing set} \\
 & Prompts & Images & Prompts & Images \\\hline \hline
Subset100K & 139,275  & 100,105 & 14,209  & 10,000  \\ \hline
Subset200K & 285,424   & 200,105 & 28,770  & 20,000   \\ \hline
Subset1M   & 1,434,563 & 940,042 & 159,245 & 104,449 \\   \hline
\end{tabular}
}
\caption{An Overview of Different Subsets. We present detailed prompts and images distributions for three subsets splited based on image quantity.}
\label{split_num}
\end{table}

\noindent \textbf{Evaluation Metrics.} The metrics used for  evaluation in Fashion-Diffusion Benchmark include:

\noindent\textbf{FID.} We evaluate generative performance using Fr\'echet Inception Distance (FID)~\cite{heusel2017gans}, a metric that computes the Fr\'echet distance between the Gaussian distributions of the SD model-generated and ground truth images.

\noindent\textbf{IS.} The Inception Score (IS)~\cite{salimans2016improved} uses the Inception model~\cite{szegedy2016rethinking} to obtain the conditional label distribution, calculates the KL-divergence between this distribution and each image's label distribution to ensure diversity, and finally exponentiates the expected divergences.

\noindent\textbf{CLIPScore.} We use CLIPscore~\cite{hessel2021clipscore} to calculate the cosine distance between the visual embedding of generated image by SD and the textual embedding of the input prompt.

\noindent\textbf{Attribute Precision.} We employ the EfficientNet-B3 model~\cite{tan2019efficientnet} trained in our fashion data to classify attributes in fashion images, and calculate the classification accuracy as the Attribute Precision for each subset.

\begin{figure*}[!t]
    \centering
    \includegraphics[width=\linewidth]{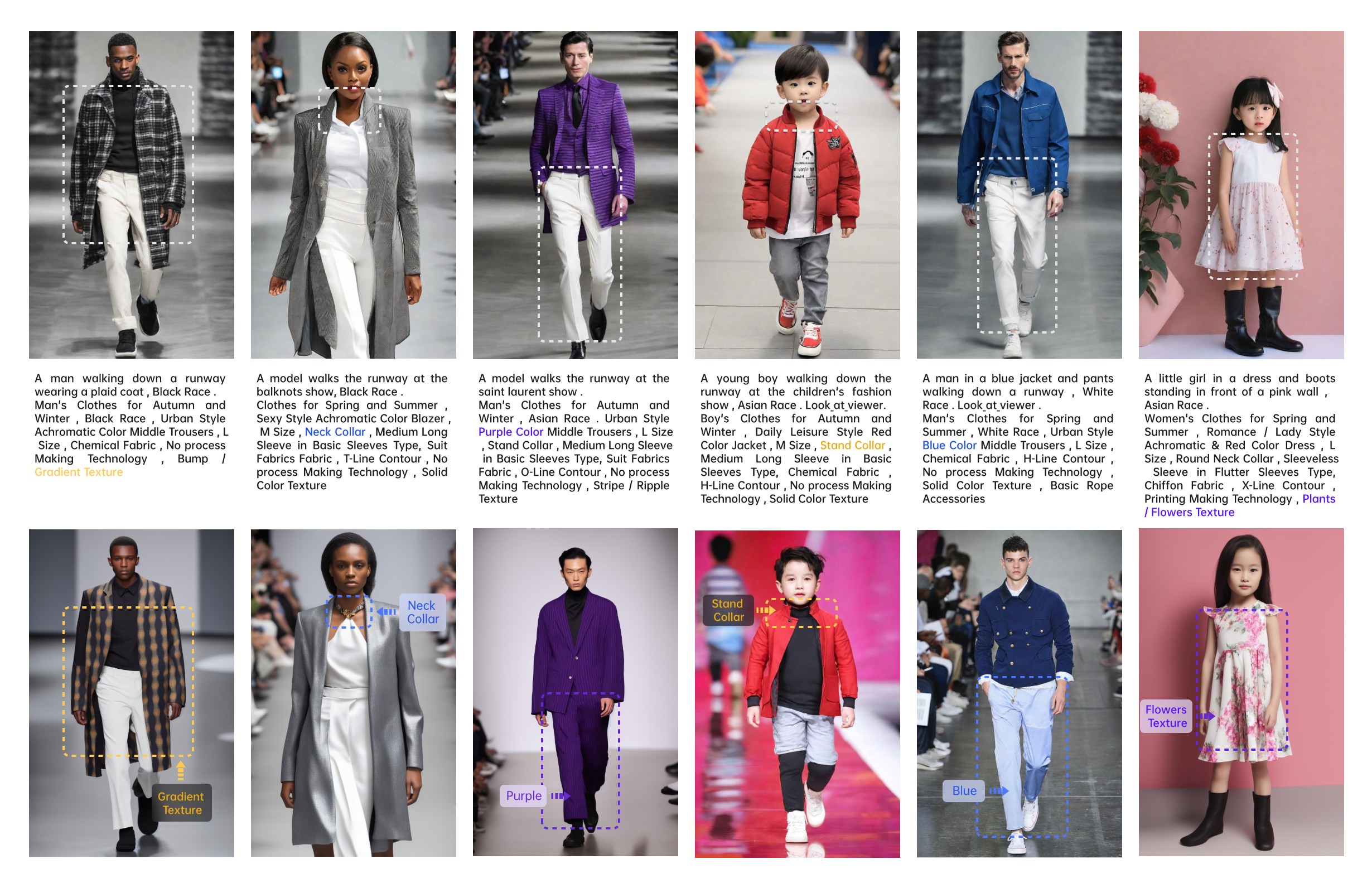}
    \caption{Qualitative Comparison. The top row shows images from the pretrained SD models, marked by significant distortions, while the bottom row presents images from SD models finetuned on Fashion-Diffusion. We annotated noticeable differences in the images, showing that our generated images better match the prompt.}
    \label{fig:quality}
\end{figure*}

\subsection{Generation Results on Fashion-Diffusion} 

\noindent\textbf{Baselines.} 
We evaluate the performance of current T2I models on the Fashion-Diffusion dataset to explore the challenges for garment synthesis. We choose the widely recognized models, \emph{e.g.} Stable Diffusion~\cite{Rombach_2022_CVPR}, for evaluation. 

\begin{table}
    \centering
    \footnotesize
    \scalebox{.92}{
    \begin{tabular}{|c|c|c|c|c|c|c|c|c|}
    \hline
\multirow{2}{*}{Models} & \multirow{2}{*}{Subsets} & \multirow{2}{*}{FID$\downarrow$} & \multirow{2}{*}{IS$\uparrow$} & \multicolumn{5}{c|}{Attribute Precision $\uparrow$}\\ \cline{5-9}
 &  &  && Category & Style&Cloth\_len&Fabric&Texture\\ \hline \hline
\multirow{3}{*}{SD-1.5} &100K & 23.76/52.02  & 6.72/5.28 & 0.38/0.34&0.19/0.15&0.32/0.31&0.24/0.19 & 0.36/0.35 \\ 
&{200K} & 24.58/49.99 & 7.02/5.38 & 0.58/0.49&0.28/0.19&0.50/0.36&0.38/0.27 &0.53/0.44   \\ 
&1M & 18.57/47.91 &  7.29/5.51 & 0.76/0.64&0.35/0.23&0.64/0.41&0.53/0.35 &0.71/0.55 \\ \hline      
\multirow{3}{*}{SD-2.1}  &100K & 17.53/35.59  & 6.26/5.75 & 0.37/0.31&0.21/0.15&0.32/0.27&0.24/0.17&0.36/0.31 \\ 
&{200K} & 12.84/33.41  & 6.64/5.74 & 0.59/0.46 &0.31/0.20&0.54/0.34 &0.41/0.24 &0.55/0.42 \\ 
&1M & 12.74/31.74 &  \textbf{7.32}/6.01 & 0.78/0.60 & 0.38/0.25& 0.69/0.40& \textbf{0.57}/0.33 &0.73/0.52 \\ \hline 
\multirow{3}{*}{SDXL}  &100K & 12.52/41.30  & 6.13/5.39 & 0.37/0.36 & 0.20/0.17 & 0.31/0.29 & 0.22/0.20 & 0.33/0.34 \\ 
&{200K} & 9.13/38.08 &  6.51/5.53&  0.58/0.52 & 0.30/0.23 & 0.54/0.39 & 0.38/0.30 & 0.53/0.46\\ 
&1M & \textbf{8.33}/36.94 & 6.95/5.72& \textbf{0.78}/0.69& \textbf{0.40}/0.28& \textbf{0.69}/0.48 & 0.56/0.41 & \textbf{0.74}/0.58  \\ \hline
    \end{tabular}}
    \caption{Comparisons of SD models trained on three different splitting levels of Fashion-Diffusion Dataset. We achieve the continuous improvements result can on all models when training and evaluating on our three subsets. For clarity in comparison, we present all results in the format of \textit{Finetuned/Pretrained}.} 
    \label{tab:splitted}
\end{table}

\noindent\textbf{Results on different subsets.} 
We assess SD models across three levels in Fashion-Diffusion, \emph{i.e.} Subset100K, Subset200K, and Subset1M. Substantial results in Table~\ref{tab:splitted} showcase that training on more data can continually improve the performances of the generative models. It clearly exhibits a decreasing trend on FID and an increasing trend on both IS and Attribute Precision for all SD series models. For example, SDXL finetuned on Subset100K obtains 12.52 FID, and gains to 9.13 FID by finetuning on Subset200K, and achieves 8.33 FID (SOTA in Fashion-Diffusion) after finetuning on Subset1M.
To assess the capability of generating fine-grained attributes, we intuitively compare the Attribute Precision of the images generated by finetuned and pre-trained models on the five classes, i.e. `Category', `Style', `Cloth\_len', `Fabric' and `Texture', Interestingly, SD models finetuned on our subsets can boost all the results in terms of Attribute Precision.


\noindent\textbf{Comparisons on different models.} 
We finetune various top T2I models (SD-1.5, SD-2.1, SDXL) on our Fashion-Diffusion dataset to broaden evaluation. Results (Table~\ref{tab:splitted}) show SDXL's notable gains (4.19\% in FID, 0.82\% in IS) when trained on our data showcasing our dataset's efficacy in enhancing T2I models.

\noindent\textbf{Qualitative Results.} 
As shown in Fig.~\ref{fig:quality}, SD models finetuned on our dataset can generate accurate clothing and humans (bottom row) that correspond closely with prompts, compared with pretrained ones (top row). For instance, in the second column, pretrained SD can not generate a woman wearing a 'Neck Collar', while finetuned SD can do it correctly. 
Notably, our images exhibit more realistic faces, appropriately shaped bodies, and correct finger counts.

\subsection{Comparison of Generation Results on Different Datasets}

\begin{figure}[!htb]
    \centering
    \includegraphics[width=.9\textwidth]{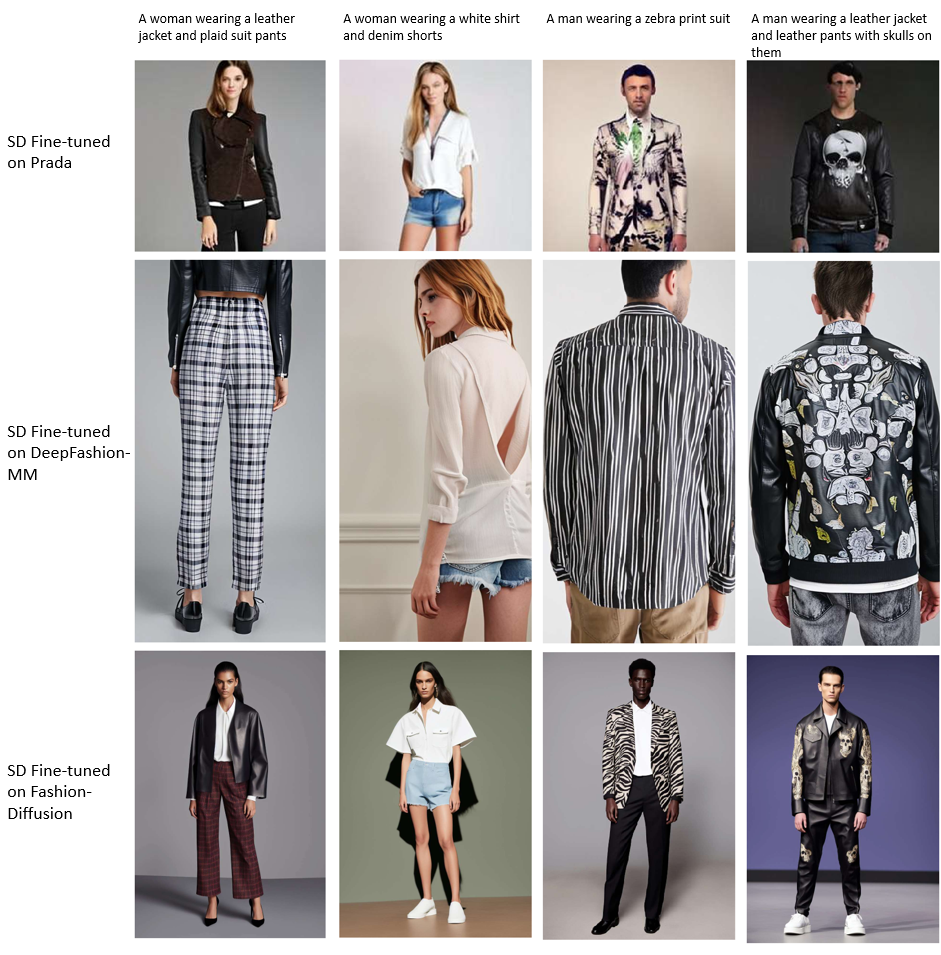}
    \caption{Generation comparisons of SDXL fine-tuned on different datasets. Fine-tuning on our Fashion-Diffuison dataset yields more accurate generation results that are better aligned with the input textual description.}
    \label{fig: appen 3}
\end{figure}

For comparison, we select Prada~\cite{zhu2017your} and DeepFashion-MM~\cite{jiang2022text2human} as baseline datasets. To ensure a fair comparison, we require the datasets to include images comprising both garments and humans, paired with detailed text descriptions. Prada and DeepFashion-MM are the only two of this kind that are publicly accessible. Table~\ref{tab:base_datasets} reports the comparison results using different SDXL models, fine-tuned on Prada, DeepFashion-MM, and Fashion-Diffusion, for generation. 
Based on the FID, IS, and CLIPScore comparisons, we observe that our dataset yields the best generation results, with FID 8.33, IS 6.95, and CLIPScore 0.83.
In addition, some qualitative results are shown in Fig.~\ref{fig: appen 3}. SD fine-tuned on our dataset can generate images that are better aligned with textual description, compared to SD fine-tuned on Prada and DeepFashion-MM.

\begin{table}
    \centering
    \scalebox{1}{
    \begin{tabular}{|c|c|c|c|c|}
    \hline
        Dataset  &FID $\downarrow$ & IS$\uparrow$ & CLIPScore$\uparrow$ \\ \hline
        \hline
        Prada~\cite{zhu2017your}& 18.36 & 4.23 & 0.70  \\ \hline
        DeepFashion-MM~\cite{jiang2022text2human}& 15.32 &4.72&0.70\\
        \hline
        \textbf{Fashion-Diffusion} & \textbf{8.33} & \textbf{6.95} & \textbf{0.83} \\ \hline
    \end{tabular}
    }
    \caption{Comparisons of different datasets. We use SDXL as the base model and compare FID, IS and CLIPScore on three different fashion datasets.}
    \label{tab:base_datasets}
\end{table}
\vspace{-5mm}

%% file: sec/6_con.tex
\section{Conclusion}
\label{sec:con}
This paper introduces and assesses the Fashion-Diffusion dataset, the first to offer over a million images for T2I-based fashion design research. Its extensive collection of high-quality human-garment pairs and detailed clothing attributes promises to spur advancements in fashion design. Statistical analysis confirms its text and image quality, and text-image relevance, making it a dependable resource for future studies.

We've also established a new benchmark from the Fashion-Diffusion dataset for standardization in the fashion design field, which enhances consistency and comparability across different models, thereby fast-tracking innovation.

Plans are underway to expand the dataset and use its unique human-related data for human image generation, potentially paving the way for applications in virtual try-ons, fashion design, and virtual reality. In essence, the Fashion-Diffusion dataset marks a significant leap in fashion technology, offering new pathways for T2I-based fashion design research and development.

%% file: sec/X_suppl.tex
\clearpage
\centerline{\textbf{\large{Appendix}}}
\appendix

\section{Detailed annotations for attributes}
\label{sec:attri}
We engage fashion design professionals to categorize subjects into 23 clothing design classes (Table \ref{tab:attri}, column 1). Each class includes diverse attributes, with the count detailed in column 2, alongside specific examples. In total, 8037 attributes comprehensively describe the clothing subjects. The right portion of Table \ref{tab:attri} provides the size of manually annotated data subsets, models to be trained, and the prediction accuracy on the validation set.

\begin{table}[!hb]
    \centering
    \scriptsize 
    \begin{tabular}{|p{.12\columnwidth}|p{.08\columnwidth}|p{.42\columnwidth}|p{.06\columnwidth}|p{.18\columnwidth}|p{.1\columnwidth}|}
    \hline
        \multirow{2}{*}{Class} & \multicolumn{2}{c|}{Attributes}   & \multicolumn{3}{c|}{Manual Annotations (Privacy)}\\ \cline{2-6}
         &  Number & Examples & Size & Model & Accuracy\\ \hline \hline
        gender & 2 & Women's clothing, men's clothing & 100K & EfficientNet-B3 & 0.98\\ \hline
        season & 2 & Spring and Summer, Autumn and Winter & 20K & EfficientNet-B3 & 0.95\\ \hline
        collar & 21 & Lapel collar, stand-up collar, etc. & 50K & EfficientNet-B3 & 0.87\\ \hline
        sleeve & 3 & Medium long, Sleeveless, Short & 20K & EfficientNet-B3 & 0.95\\ \hline
        sleeve type & 20 & Patchwork sleeve, Fur sleeves,  etc & 80K & EfficientNet-B3 & 0.83  \\ \hline
        fabric & 26 & Formal fabric, Woolen fabric, etc & 100K & EfficientNet-B3 & 0.80\\ \hline
        contour & 5 & H-type, X-shaped, S-type, O-type, T-shaped & 20K & EfficientNet-B3 & 0.79\\ \hline
        clothes length & 3 & Long, Medium, Short & 50K & EfficientNet-B3 & 0.88\\ \hline
        style & 25 & Athletic, Luxury, Loungewear, Lolita, etc & 150K & EfficientNet-B3 & 0.83\\ \hline
        garment category & 52 & Fur Coat, Backless Pants, Denim Shirt, etc & 400K & EfficientNet-B3 & 0.89\\ \hline
        technology & 39 & Fine Stitch, Knitted Threads, Printing, etc & 80K & EfficientNet-B3 & 0.76\\ \hline
        texture & 33 & cartoon sub, swoosh, diamond, floral, etc & 100K & EfficientNet-B3 & 0.78\\ \hline
        accessories & 24 & Decorative Zippers, Sequins , Fringes, etc & 100K & EfficientNet-B3 & 0.73\\ \hline
        look-at-view & 2 &True, False & 10K & ResNet-50 & 0.85 \\ \hline
        view & 5 & Close, Upper, Mid-length, Full, Other & 10K & ResNet-50 & 0.87 \\ \hline
        weight & 2 & Fat, Thin &  &  &  \\ \hline
        complexion & 5 & White, Yellow, Brown, Black, Other & 4K & CLIP+MLP & 0.91\footnotemark[1] \\ \hline
        color-3 & 904 & Chicory coffee, Teal Blue, Peach White, etc.& 1M &EfficientNet-B1\footnotemark[2] & 0.90 \\ \hline
        color-2 & 268 & Palace Blue, Light Mint Green, Light Gold, etc.& 1M &\multirow{2}{*}{\parbox{.18\columnwidth}{Aggregated by fine-grained color-3 attributes}} & \multirow{2}{*}{-} \\ \cline{1-4}
        color-1 & 9 & Pink, Red, Orange, Yellow, etc.& 1M & & \\ \hline
        color & 8 & Red, Orange, Yellow, Green, Blue, etc. & - & LeNet + KNN & - \\ \hline
        location & 149 & Milan, Madrid, Tokyo, New York, Berlin, etc.& 1M &Extract from Runway title & - \\ \hline
        brand & 6430 & Holiday, Maison Anoufa, Amber Holmes, Harman Grubisa, etc.& 1M &Extract from Runway title & - \\ \hline        
    \end{tabular}
    \caption{\footnotesize Detailed attributes and manual annotations. Initially, we use manually annotated data to train attribute detection models. Then we use trained models to label the extra large data. For clear visualization, we organize it in three parts, i.e. ``Class", ``Attributes" and ``Manual Annotations".}
    \vspace{-4mm}
    \label{tab:attri}
\end{table}

\footnotetext[1]{We achieve 0.91 \textit{Accuracy} in attributes of `White', `Yellow', and `Black'. We use \textit{Recall} metric for evaluating the `Brown' attribute, obtaining 0.90 score.}
\footnotetext[2]{We utilize an unsupervised methodology to train the EfficientNet-B1 model, which yields a top-3 accuracy rate of 90\%.}

\section{Aesthetic quality comparisons}
\begin{figure}[!hbt]
    \centering
    \includegraphics[width=.96\linewidth]{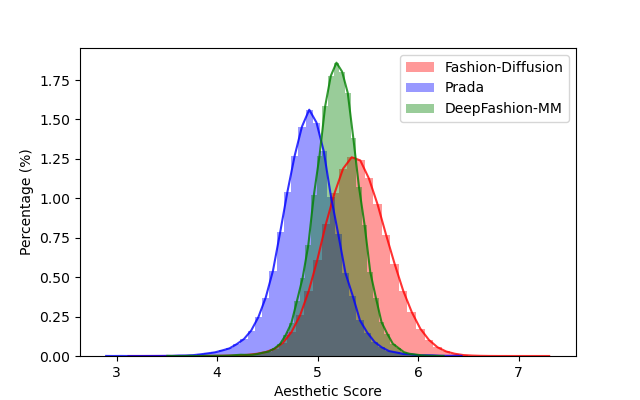}
    \caption{\small Aesthetic image quality comparisons between different datasets, i.e. Fashion-Diffusion (ours), Prada and DeepFashion-MM. Evidently, our dataset of 1.04 million fashion images has yielded the highest aesthetic score, which is a testament to the superior quality of the images we have curated.}
    \label{fig:athesic}
\end{figure}

We analyze the quality of our collected Fashion-Diffusion dataset. To demonstrate its superiority, we compare with two other fashion datasets, i.e Prada and DeepFashion-MM. We use  LAION Aesthetics Predictor V2\footnotemark to calculate the Aesthetic Score for evaluating the quality of fashion images. The aesthetic quality of all datasets is displayed in Fig.~\ref{fig:athesic}. Fashion-Diffusion attains a mean Aesthetic Score of 5.38, outperforming Prada's 4.91 and DeepFashion-MM's 5.19. This signifies Fashion-Diffusion's superior quality for fashion design.

\footnotetext{https://github.com/christophschuhmann/improved-aesthetic-predictor}

\begin{table}[!htb]
    \centering
    \footnotesize
    \scalebox{.91}{
    \begin{tabular}{|c|c|c|c|c|c|c|}
    \hline
\multirow{2}{*}{Models}  &\multirow{2}{*}{Subsets} & \multicolumn{5}{c|}{Attribute Precision $\uparrow$}\\  \cline{3-7}
 & & accessories & collar &technology &color-1  & sleeve type  \\ \hline \hline
\multirow{3}{*}{SD-1.5}& 100K&0.12/0.08 &0.14/0.07& 0.30/0.26& 0.22/0.23&0.22/0.22 \\ 
 &200K&0.17/0.11 &0.22/0.10& 0.44/0.34& 0.31/0.34 &0.30/0.28 \\ 
 &1M&0.25/0.14&0.32/0.15&0.58/0.41&0.43/0.45&0.42/0.36\\ \hline
\multirow{3}{*}{SD-2.1}&100K&0.11/0.07 &0.14/0.08& 0.30/0.24& 0.25/0.19 & 0.20/0.21  \\ 
 &200K&0.19/0.11 &0.24/0.12& 0.48/0.31& 0.43/0.28 & 0.33/0.29   \\ 
 &1M&0.27/0.14 &0.35/0.18& 0.64/0.39& 0.60/0.37 & 0.46/0.38   \\ \hline
\multirow{3}{*}{SDXL}& 100K&0.11/0.10 &0.13/0.12& 0.28/0.25& 0.23/0.17 & 0.21/0.25  \\ 
 &200K&0.19/0.15 &0.23/0.19& 0.44/0.34& 0.40/0.26 & 0.34/0.34   \\ 
 &1M&\textbf{0.29}/0.20 &\textbf{0.36}/0.28& \textbf{0.66}/0.43& \textbf{0.62}/0.35& \textbf{0.46}/0.43   \\ \hline
    \end{tabular}}
    \caption{More Attribute Precision Results in Fashion-Diffusion. We achieve the continuous improvements result in terms of Attribute Precision in additional classes, i.e. ``accessories", ``collar", ``technology", ``color-1" and ``sleeve type" on all models when training and evaluating on our three subsets. Similar as in the paper, we present all results in the format of \textit{Finetuned/Pretrained}.}
    \label{tab:AP_more}
\end{table}

\section{More Attribute Precision Results}
\label{sec:diff_res}
We further elaborate on our findings related to Attribute Precision for various classes, building upon the data presented in Table~\ref{tab:splitted} in the paper. A comprehensive analysis of the results in Table~\ref{tab:AP_more} reveals that the fine-tuned models consistently outperform the pre-trained models across all subsets. This observation underscores the effectiveness of our dataset in enhancing model performance.

Moreover, the model SDXL exhibits the highest Attribute Precision for several classes, including ``accessories", ``collar", ``technology", ``color-1" and ``sleeve type". This highlights the model's proficiency in accurately identifying these specific attributes.

\section{Fashion design tool}
\label{sec:paltform}
\begin{figure}[!hbt]
    \centering
    \includegraphics[width=\linewidth]{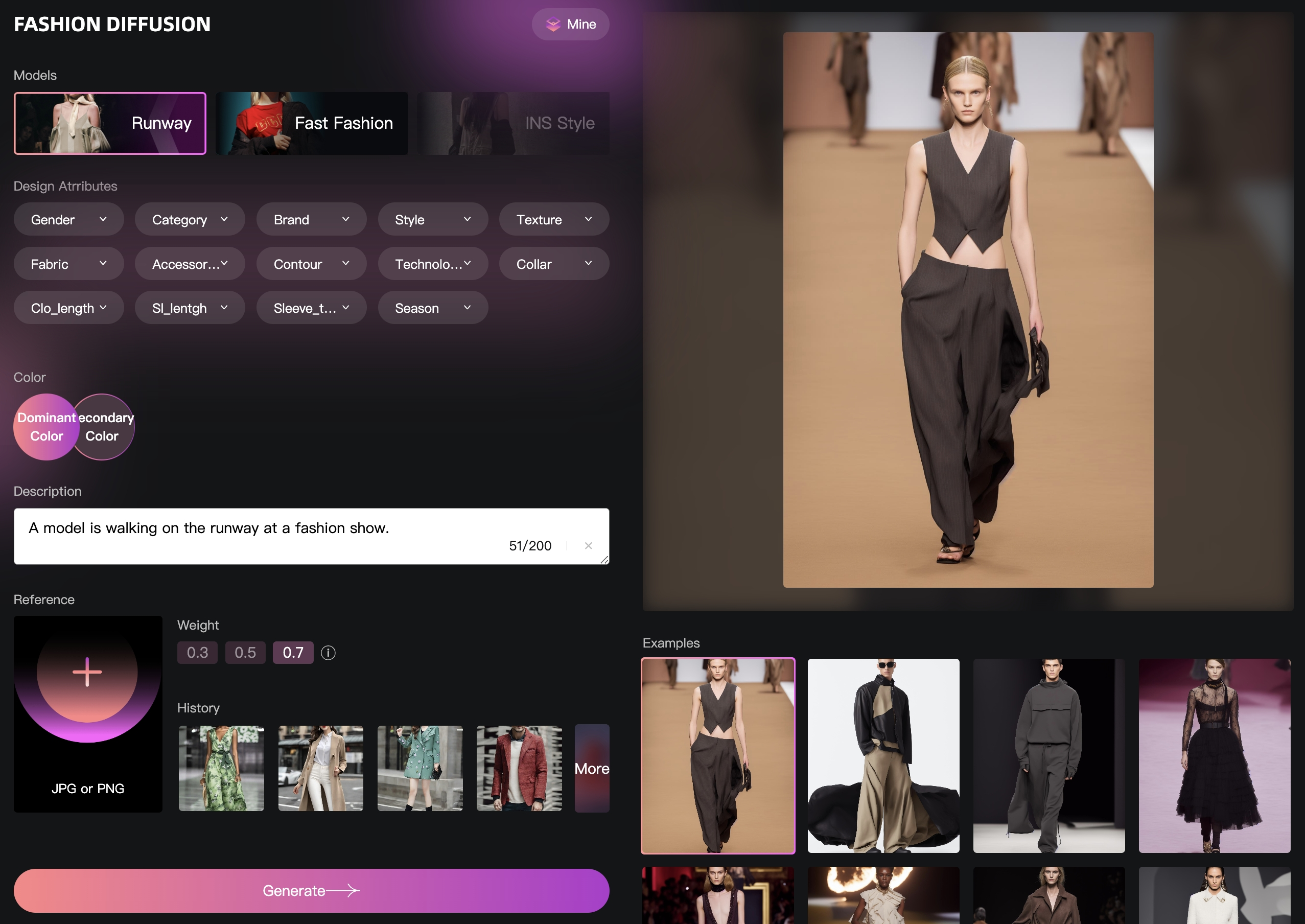}
    \caption{\small Fashion design tool based on Fashion-Diffusion.}
    \label{fig:platform}
\end{figure}

We have developed a tool Fig.~\ref{fig:platform} for fashion design, which is fundamentally based on the principles of Fashion-Diffusion. This tool leverages the insights and methodologies of Fashion-Diffusion to provide a robust and intuitive platform for creating and analyzing fashion designs. This tool further highlights the necessity and utility of fine-grained attributes. It demonstrates how, by selecting models, colors, design attributes, and weights, we can create diverse fashion images. Essentially, it shows that fine-grained attributes enable the simulation of various fashion styles on chosen models.

\section{Filtering rules}
\label{sec:filtering_rules}
We initiate the process by sampling high-quality, diverse fashion images from our raw collections. This is followed by a pre-processing filtering stage to refine the dataset, resulting in three distinct subsets: Subset100K, Subset200K, and Subset1M.  

Our filtering approach is nuanced, considering various factors like scale factor, garment features, human characteristics, image attributes, and specific attribute constraints. These considerations help us meticulously filter the datasets during the subdivision process. We have created a bespoke filtering procedure, encompassing five filtering rules, as detailed in Table \ref{tab:filtering_rule}.

\begin{table}[!htb]
    \centering
    \begin{tabular}{|p{.17\columnwidth}|p{.37\columnwidth}|p{.42\columnwidth}|}
    \hline
        Filtering rules & Filtered out & Retained \\ \hline
        Scale rule & & aspect ratio==0.67, width>=768  \\ \hline
        Clothing rule & Empty in any of \{`garment category', `color', `gender', `season', `technology', `texture'\}, clothes\_length is N/A, garment category is N/A or `boot' & gender in \{`Men's clothing', `Women's clothing'\} \\ \hline
        Human rule & face\_col==4, view in \{0,4\} & face\_count==1 \\ \hline
        Image rule & aesthetic<=5, cv\_var<=100 & 15<cv\_lightness<240, 20<cv\_saturation<150 \\ \hline
        Specific cases&description of sleeve\_type & view==3, clothes\_length in \{`Long', `Medium', `Short'\}, contour in \{`H-type', `A', `Normal', `X-shaped', `S-type', `O-type', `T-shaped'\}, sleeve in \{`Medium long sleeves', `Sleeveless', `Short sleeved'\} \\\hline
    \end{tabular}
    \caption{We construct the three-level subsets by using strict filtering constraints, concluded as five filtering rules, i.e. `Scale rule', `Clothing rule', `Human rule', `Image rule', and `Specific cases'.}
    \label{tab:filtering_rule}
\end{table}

\section{Controllable generation compared with original SD}
\begin{figure}[!h]
    \centering
    \includegraphics[width=.9\textwidth]{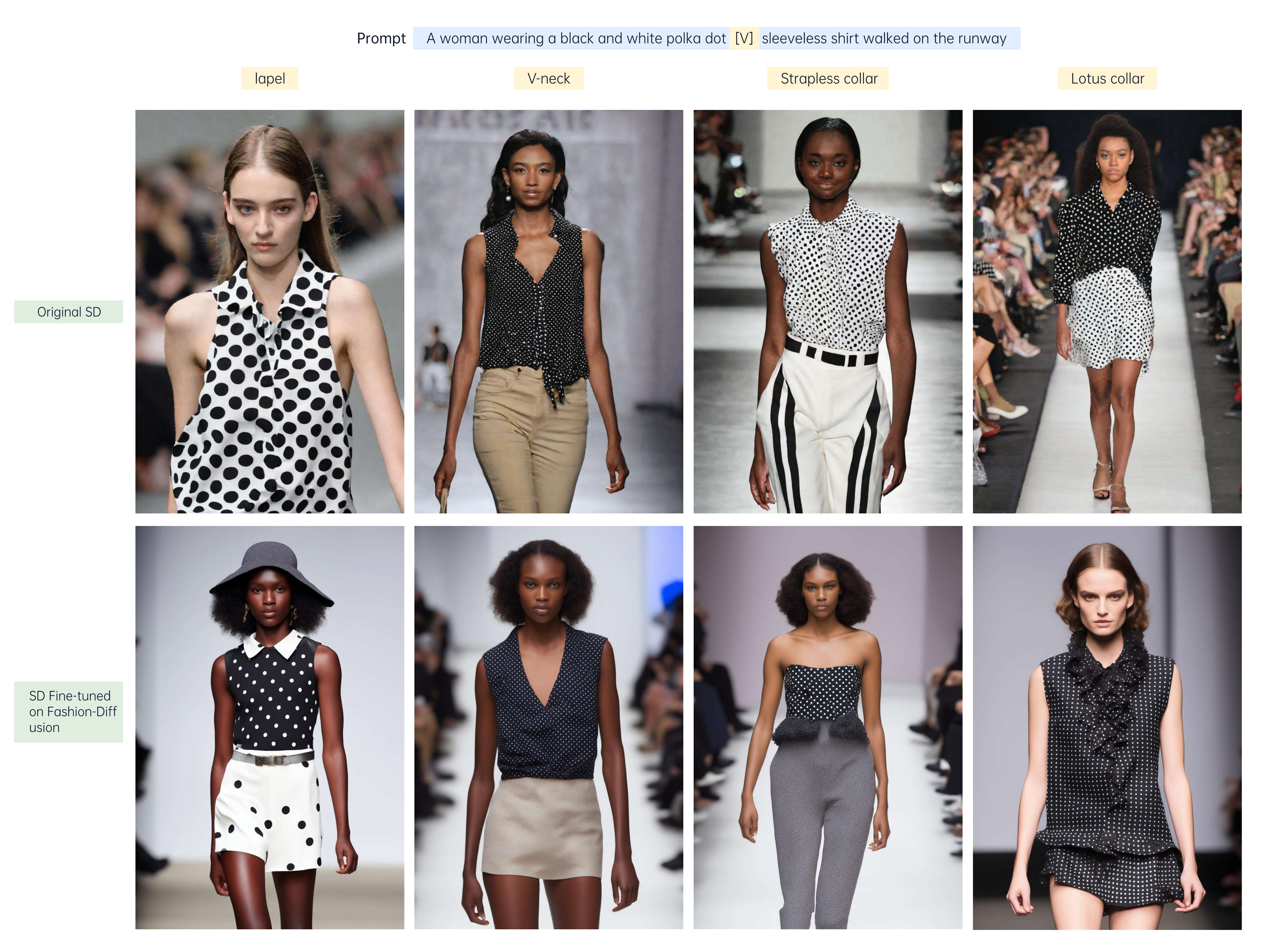}
    \caption{Generation comparisons between the original model and models trained on Fashion-Diffusion dataset. With the prompt ``A woman wearing a black and white polka dot sleeveless shirt walked on the runway", we test several different collars, e.g. ``lapel", ``V-neck", ``strapless collar", and ``Lotus collar". We can see that in ``strapless collar", our female model is exactly off-the-shoulder collar, comparing with the lapel in original SD. 
In ``lotus collar", our model are as likely as what we prompt, but the original SD generates V-neck collar.}
    \label{fig: appen 11}
\end{figure}

\begin{figure}[!h]
    \centering
    \includegraphics[width=.9\textwidth]{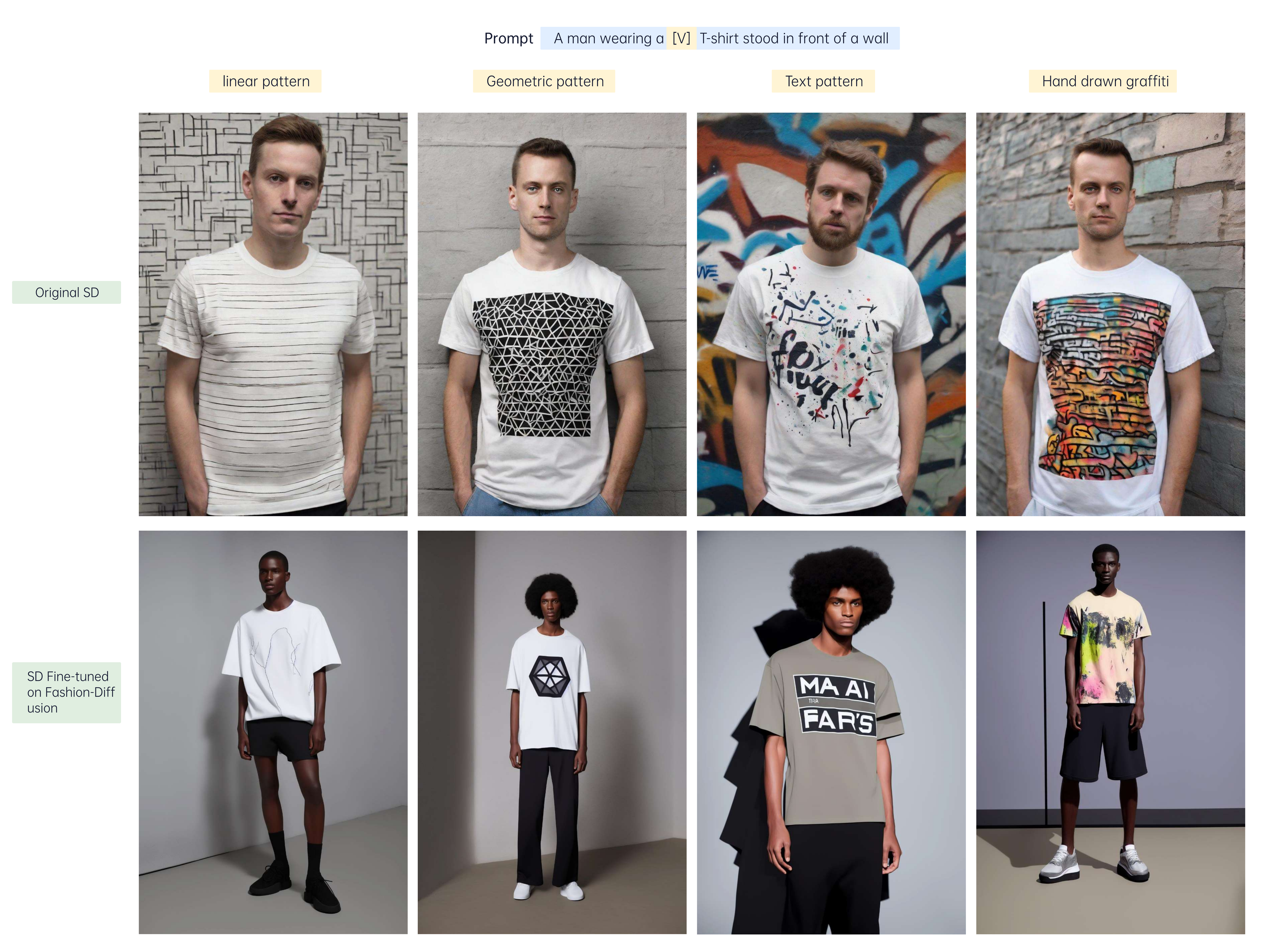}
    \caption{Generation comparisons between the original model and models trained on Fashion-Diffusion dataset. With the prompt ``A man wearing a T-shirt stood in front of a wall", we can generate artistic style and casual style, in line with the style of show models. Compared with the original SD, we can obviously generate patterns as descriptions, such as ``linear pattern", ``Geometric pattern", ``Text", etc. for further specifications.}
    \label{fig: appen 12}
\end{figure}

\begin{figure*}[]
    \centering
    \includegraphics[width=.9\textwidth]{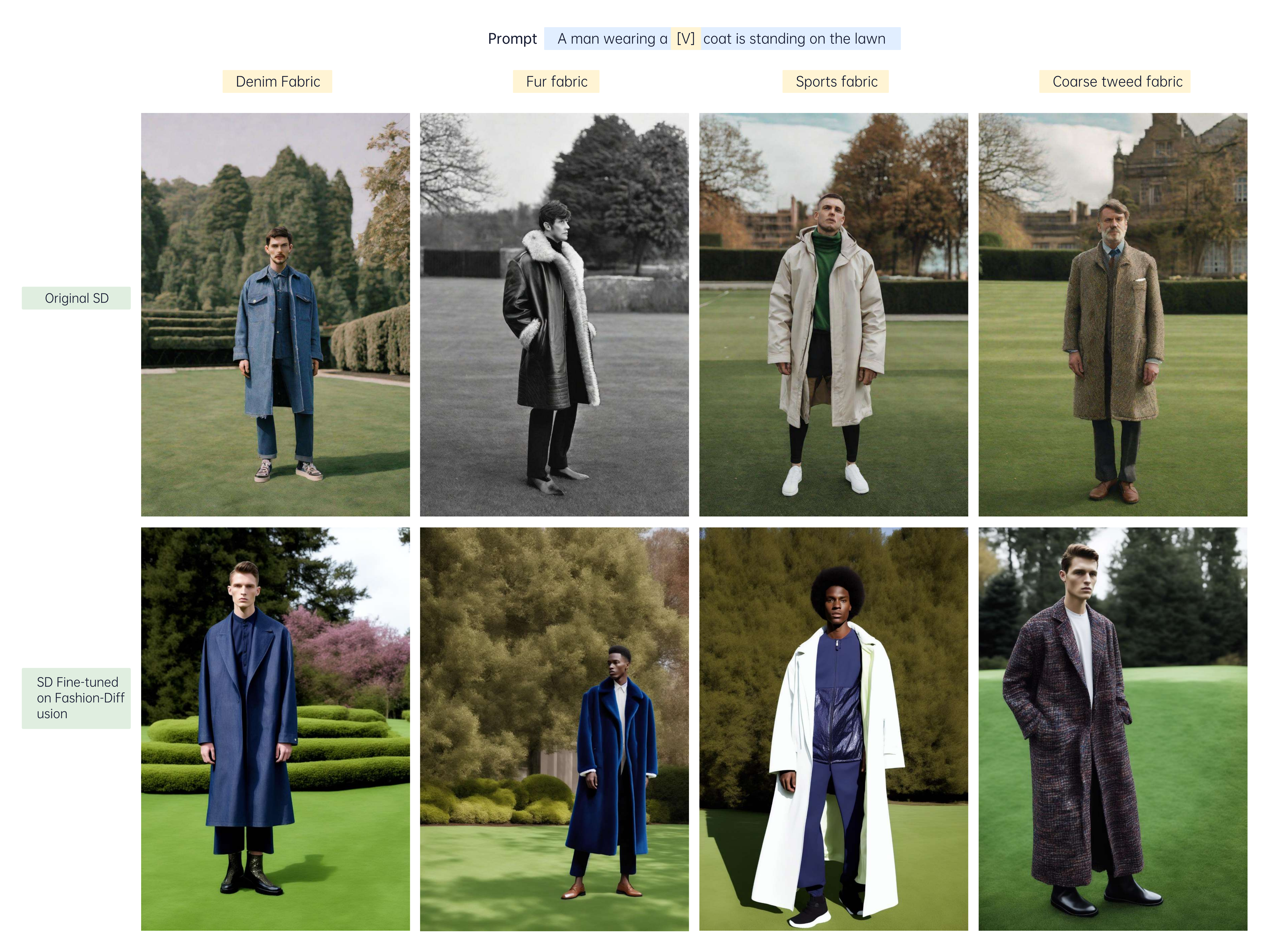}
    \caption{Generation comparisons between the original model and models trained on Fashion-Diffusion dataset. Specifically, with the prompt ``A man wearing a coat is standing on the lawn", we can generate standard male models in a coat and control the model to wear specific fabrics, e.g. ``Denim", ``Fur" etc.}
    \label{fig: appen 13}
\end{figure*}

We illustrate more generation comparisons in terms of specific cloth styles, fabric, patterns, etc., by which we aim to evaluate the effectiveness of attributes in our dataset. The detailed information is presented in~\cref{fig: appen 11,fig: appen 12,fig: appen 13}, where the type is denoted using \colorbox{yellow}{[V]} and highlighted with a light yellow background, in accordance with the prompt. Clearly, the fine-tuned model can perform controllable generation based on different types, showcasing a significant improvement over the original SD.
 

\section{Generation comparisons with other datasets}
\label{sec:gen_datasets}
To clearly demonstrate the huge capacity of our dataset, we illustrate more generation comparisons qualitatively in terms of various attributes as the following figures.

\begin{figure*}[]
    \centering
    \includegraphics[width=.9\textwidth]{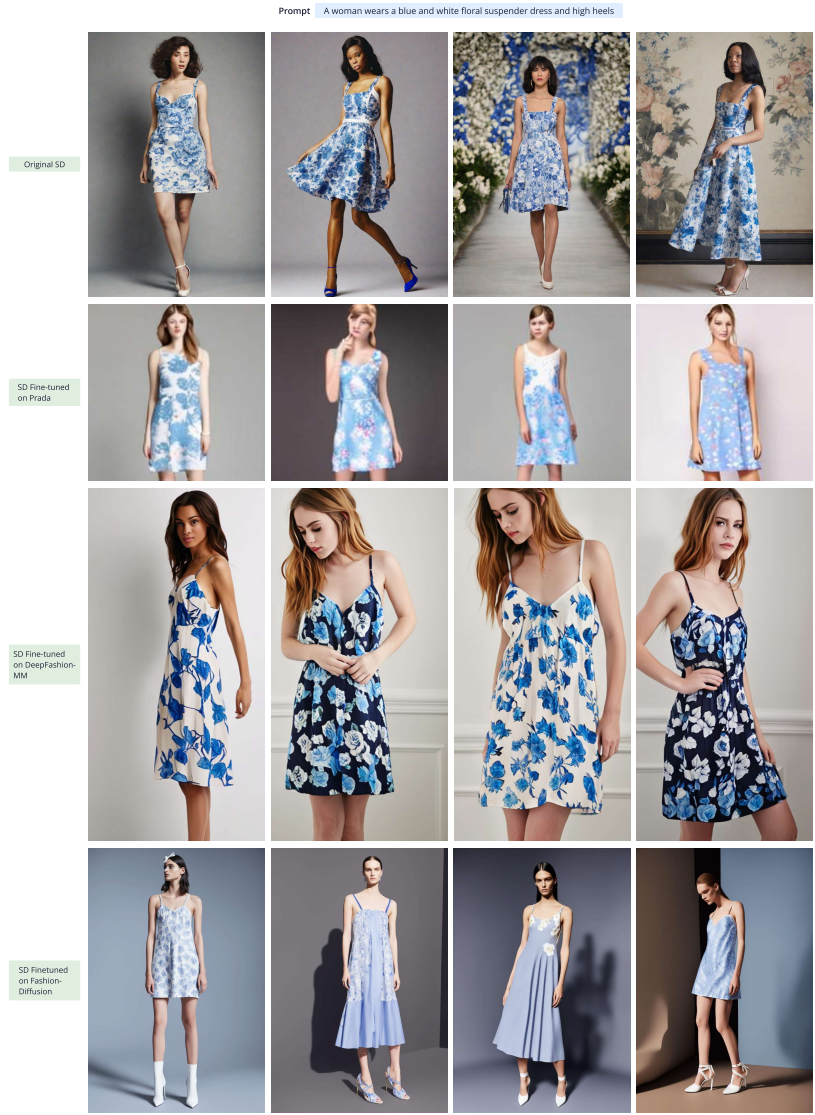}
    \caption{\small Generation comparisons with the original model and models trained other datasets. Specifically, we can generate clothes of more lifelike, more vivid, and natural by the guidance of ``A woman wears a blue and white floral suspender dress and high heels".}
    \label{fig: appen 1}
\end{figure*}

In Figure~\ref{fig: appen 1}, we prompt the original SD model and fine-tuned SD model on our Fashion-Diffusion dataset to implement the text guide, i.e. ``A woman wears a blue and white floral suspender dress and high heels". From the results, we can see that the four models generated wearing our designated clothes look more lifelike, more vivid, and natural. Specifically, the ``floral suspender dress" and the ``high heels" are generated excitedly meet what we describe. In comparison, the models generated by the original SD model look more like with fake faces and unnatural poses, and the generated clothes are far from achieving the effect of a model show.

\begin{figure*}[]
    \centering
    \includegraphics[width=.9\textwidth]{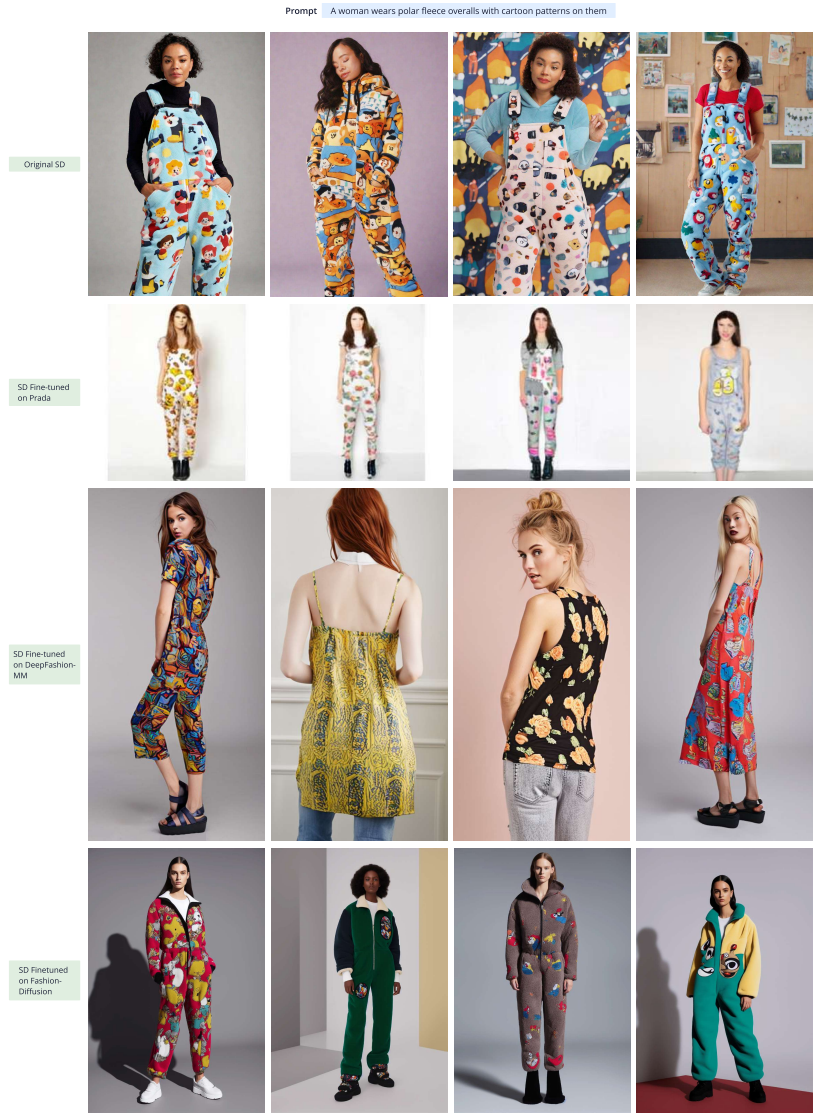}
    \caption{Generation comparisons with the original model and models trained other datasets. With the prompt of ``A woman wears polar fleece overalls with cartoon patterns on them", we can generate generous, decent and good-looking polar fleece overalls, matching the seller’s clothing display style.}
    \label{fig: appen 2}
\end{figure*}

\begin{figure*}[]
    \centering
    \includegraphics[width=.9\textwidth]{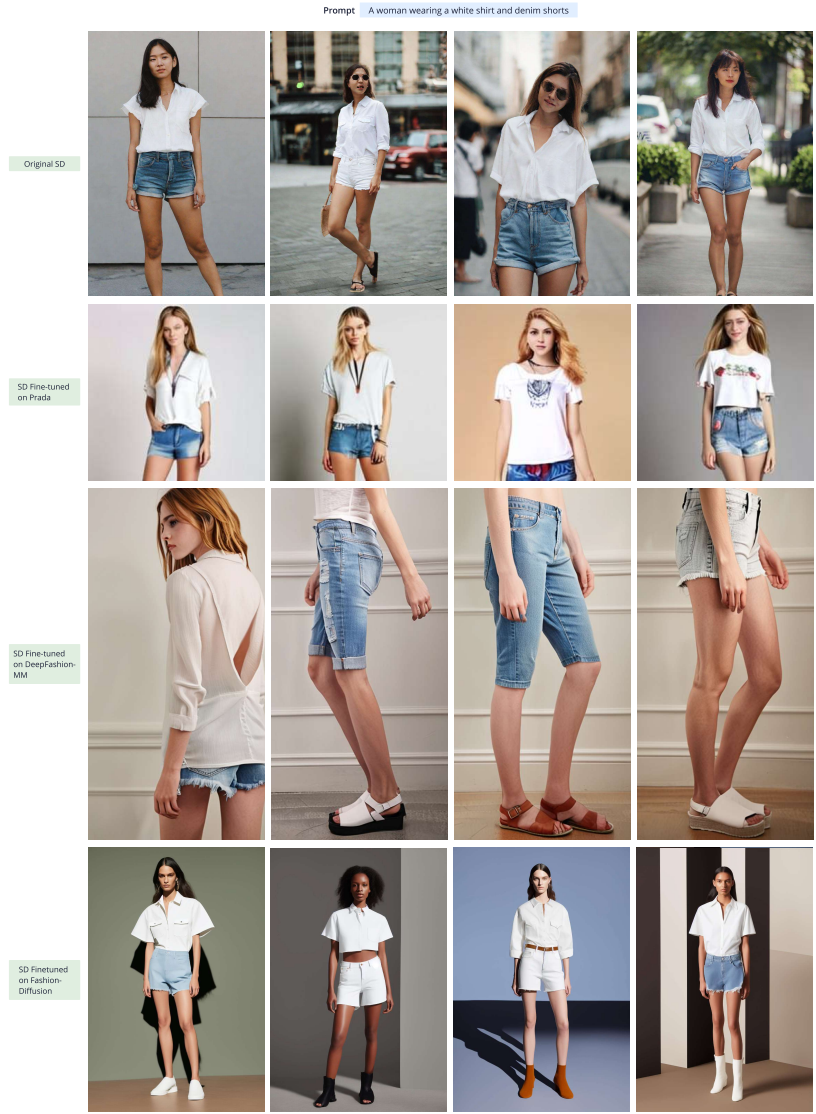}
    \caption{Generation comparisons with the original model and models trained other datasets. We can generate synthesis images with formal catwalk models, by the guidance of ``A woman wearing a white shirt and denim shorts".}
    \label{fig: appen 4}
\end{figure*}

\begin{figure*}[]
    \centering
    \includegraphics[width=.9\textwidth]{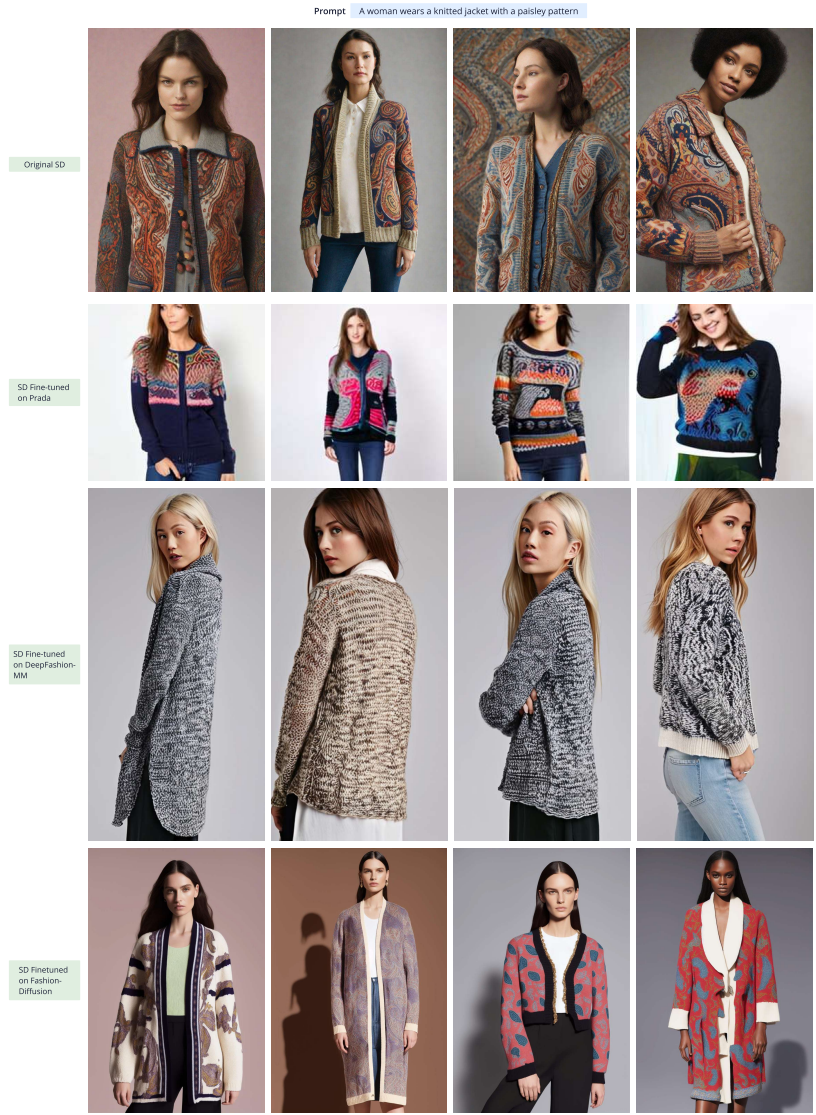}
    \caption{Generation comparisons with the original model and models trained other datasets. According to the text prompt of ``A woman wears a knitted jacket with a paisley pattern", we can generate various knitted jackets in the pattern of paisley, fully realistically without any sense of violation.}
    \label{fig: appen 5}
\end{figure*}

\begin{figure*}[]
    \centering
    \includegraphics[width=.9\textwidth]{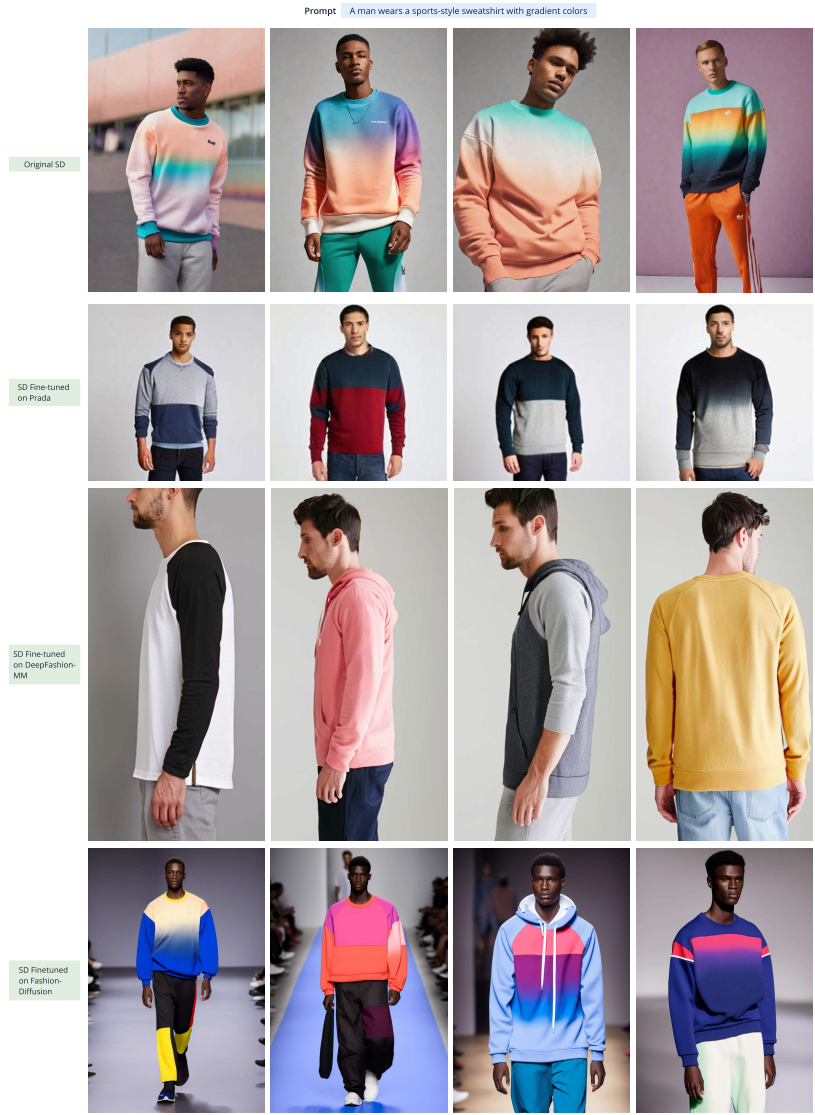}
    \caption{Generation comparisons with the original model and models trained other datasets. We can generate standard male models wearing sweatshirt with various sports-styles, with the prompt ``A man wears a sports-style sweatshirt with gradient colors".}
    \label{fig: appen 6}
\end{figure*}

\begin{figure*}[]
    \centering
    \includegraphics[width=.9\textwidth]{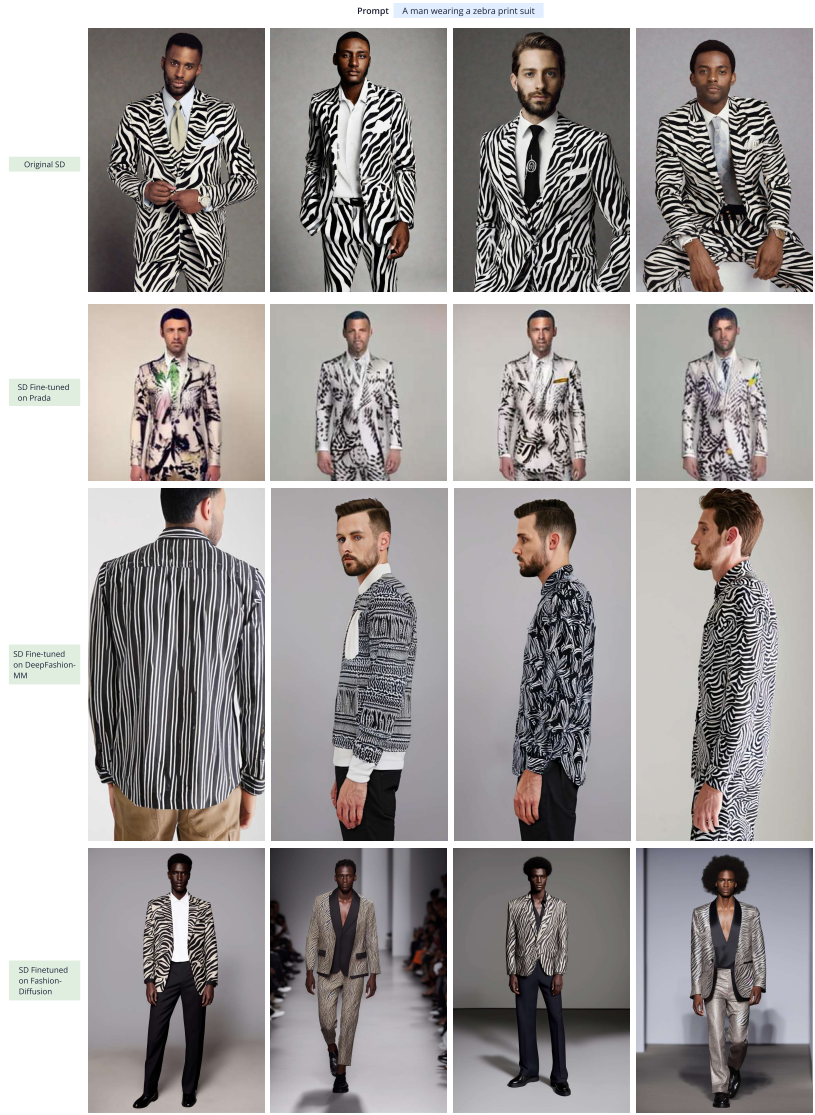}
    \caption{Generation comparisons with the original model and models trained other datasets. We can generate clothes of zebra print suit in different styles, from which the consumers can pick up in easy and comfortable experiences, by the guidance of ``A man wearing a zebra print suit".}
    \label{fig: appen 7}
\end{figure*}

\begin{figure*}[]
    \centering
    \includegraphics[width=.9\textwidth]{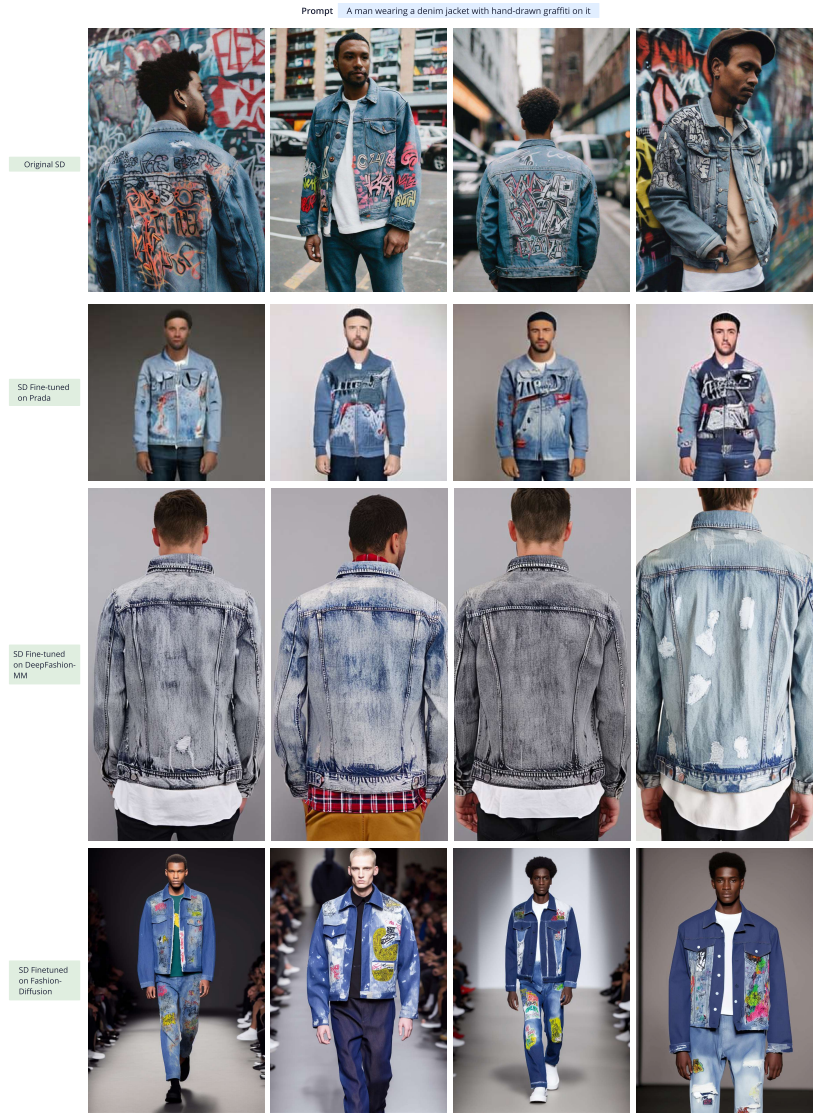}
    \caption{Generation comparisons with the original model and models trained other datasets. Specifically, by using the prompt of ``A man wearing a denim jacket with hand-drawn graffiti on it", we can generate exact male models on the catwalk. While original SD generates images aimlessly.}
    \label{fig: appen 8}
\end{figure*}

\begin{figure*}[]
    \centering
    \includegraphics[width=.9\textwidth]{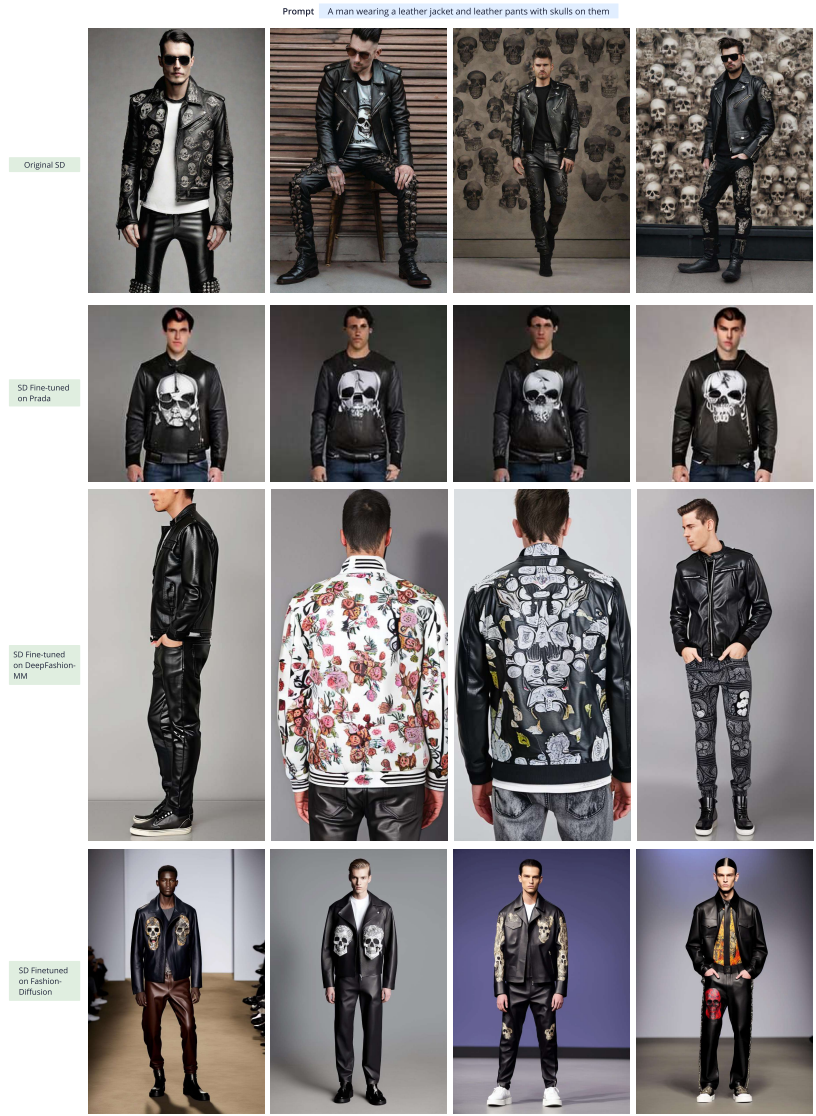}
    \caption{Generation comparisons with the original model and models trained other datasets. Specifically, using the prompt of ``A man wearing a leather jacket and leather pants with skulls on them". It shows that SD model fine-tuned by our specific huge clothing data can exactly generate clean and beautiful images, in line with style of catwalk models.}
    \label{fig: appen 9}
\end{figure*}

\begin{figure*}[]
    \centering
    \includegraphics[width=.9\textwidth]{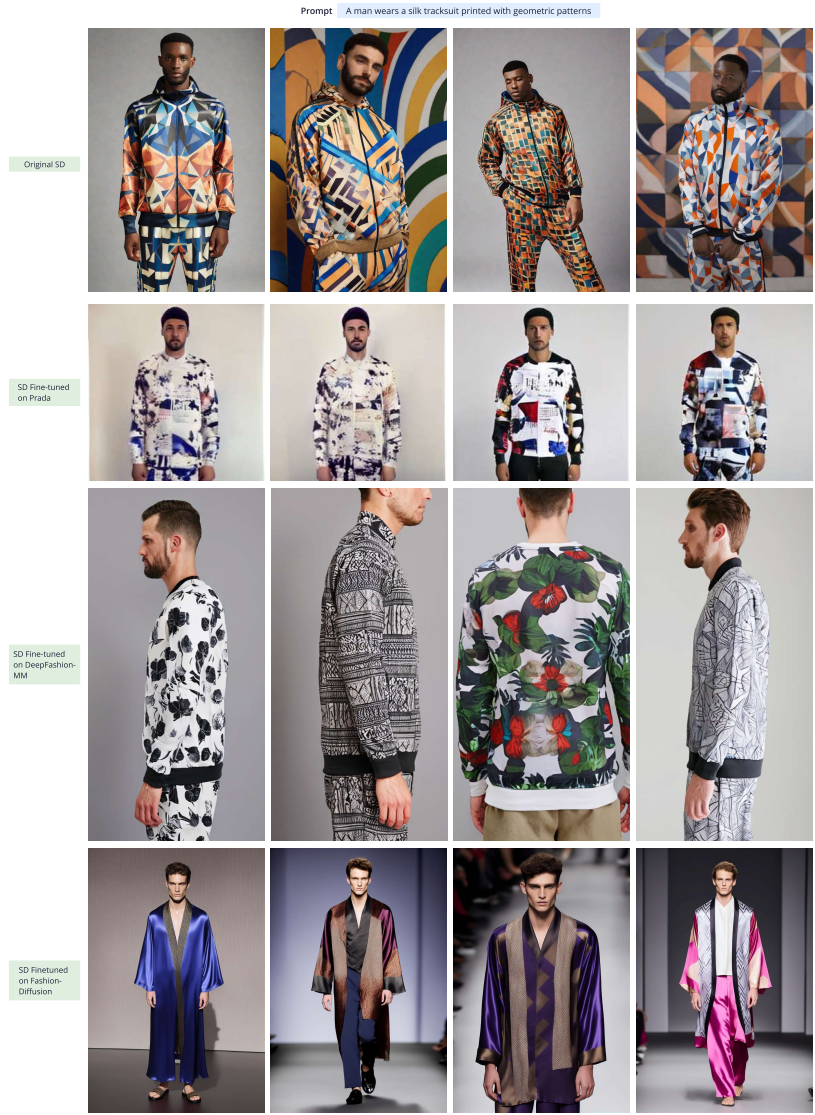}
    \caption{Generation comparisons with the original model and models trained other datasets. Specifically, by using ``A man wears a silk tracksuit printed with geometric patterns" to prompt T2I model. we can generate elegant and high-end clothes, fully satisfying the needs of the sellers.}
    \label{fig:appen10}
\end{figure*}

Additionally, we visualize more generation comparisons with the original model and models trained other datasets, e.g. ``Original SD", ``SD Fine-tuned on Prada", ``SD Fine-tuned on DeepFashion-MM" and ``SD Finetuned on Fashion-Diffusion", as in~\cref{fig: appen 2,fig: appen 4,fig: appen 5,fig: appen 6,fig: appen 7,fig: appen 8,fig: appen 9,fig:appen10}.


